\documentclass[]{fairmeta}
\usepackage{amssymb,amsmath}
\usepackage{amsthm}
\usepackage{hyperref}
\usepackage{graphicx} 
\usepackage{url}
\usepackage{amsthm}
\usepackage{thmtools}
\usepackage{mathtools}
\usepackage{upgreek}
\usepackage{dsfont}
\usepackage{cleveref}
\usepackage{xcolor}
\usepackage{bbm}
\usepackage{booktabs}
\usepackage{multirow}
\usepackage{array}
\usepackage{makecell}
\usepackage{tcolorbox}
\usepackage[table]{xcolor}
\usepackage{footnote}

\usepackage{enumitem}
\setitemize{noitemsep,topsep=0pt,parsep=0pt,partopsep=0pt}

\declaretheoremstyle[headfont=\bf,bodyfont=\normalfont]{ex}

\declaretheoremstyle[bodyfont=\normalfont]{rm}

\newcommand{\xxcomment}[4]{\textcolor{#1}{[$^{\textsc{#2}}_{\textsc{#3}}$ #4]}}
\newcommand{\julia}[1]{\xxcomment{purple}{J}{K}{#1}}
\newcommand{\tash}[1]{\xxcomment{blue}{N}{B}{#1}}
\newcommand{\TODO}[1]{\textcolor{red}{[TODO: #1]}}
\renewcommand{\julia}[1]{}
\renewcommand{\tash}[1]{}
\renewcommand{\TODO}[1]{}

\definecolor{GreedyBlue}{HTML}{003366}   
\definecolor{SampleGreen}{HTML}{006400}  
\definecolor{paleblue}{HTML}{EAF2FB}
\definecolor{paleyellow}{HTML}{FFFBEA}
\definecolor{paleviolet}{HTML}{F4EEFF}
\newcommand{\greedy}[1]{\textbf{\textit{\textcolor{GreedyBlue}{#1}}}}   
\newcommand{\sample}[1]{\textbf{\textcolor{SampleGreen}{#1}}}         

\def\R{{\mathbb{R}}}
\def\E{{\mathbb{E}}} 
\DeclareMathOperator{\Id}{Id}
\DeclareMathOperator{\softmax}{softmax}
\newcommand{\transf}{\mathcal{T}}  

\def\sg{{^\mathrm{sg}}} 
\newcommand{\norm}[1]{\left\lVert#1\right\rVert}  

\usepackage{natbib}

\title{Soft Tokens, Hard Truths}

\author[1,*]{Natasha Butt}
\author[2]{Ariel Kwiatkowski}
\author[2]{Ismail Labiad}
\author[2,3,\dagger]{Julia Kempe}
\author[2,\dagger]{Yann Ollivier}

\affiliation[1]{University of Amsterdam}
\affiliation[2]{Meta FAIR}
\affiliation[3]{New York University}

\contribution[*]{Work done during an internship at Meta}
\contribution[\dagger]{Joint senior authors}

\newcommand{\confonly}[1]{}     
  
\newif\ifarxiv  \arxivtrue
\newif\ifconference \conferencefalse

\abstract{
The use of continuous instead of discrete tokens during the Chain-of-Thought
(CoT) phase of reasoning LLMs has garnered attention recently, based on
the intuition that a continuous mixture of discrete tokens could simulate
a superposition of several reasoning paths simultaneously. Theoretical
results have formally proven that continuous tokens have much greater
expressivity and can solve specific problems more efficiently.
However, practical use of continuous tokens has been limited by strong training
difficulties: previous works either just use continuous tokens at
inference time on a pre-trained discrete-token model, or must
distill the continuous CoT from ground-truth discrete CoTs and face
computational costs that limit the CoT to very few tokens.

This is the first work introducing a scalable method to learn
continuous CoTs via reinforcement learning (RL), without
distilling from reference discrete CoTs. 
We use ``soft'' tokens: mixtures of tokens together
with noise on the input embedding to provide RL exploration.
Computational overhead is minimal, enabling us to learn continuous CoTs
with hundreds of tokens.
On math reasoning benchmarks with Llama and Qwen models up to 8B, training with continuous CoTs match discrete-token CoTs for
pass@$1$ and surpass them for pass@$32$, showing greater CoT diversity.
In systematic comparisons, 
the best-performing scenario is to train with continuous CoT tokens then use
discrete
tokens for inference, meaning the ``soft'' models can be deployed in a
standard way.
Finally, we show continuous CoT RL training better preserves the predictions of the base
model on out-of-domain tasks, thus providing a softer touch to the base
model.

}

\date{\today}
\correspondence{Natasha Butt at  n.e.butt@uva.nl}

\begin{document}

\maketitle

\section{Introduction}

Large Language Models (LLMs) have achieved impressive success across a
wide range of reasoning tasks, particularly when enhanced with
Chain-of-Thought (CoT) prompting, where models generate intermediate
``thinking tokens'' before producing final answers. While effective,
standard CoT is constrained by the discreteness of language tokens: each
intermediate step must be sampled sequentially, which can limit
expressivity and hinder exploration of diverse reasoning paths. This
contrasts sharply with human cognition, which often operates over
abstract and fluid concepts rather than rigid linguistic symbols.
Motivated by this gap, recent work has explored enabling LLMs to reason
in continuous concept spaces, a direction often termed ``continuous
CoTs'' \citep{hao2024coconut} or ``Soft Thinking''
\citep{zhang2025softthinkingunlockingreasoning}.

From a theoretical perspective, continuous reasoning offers significant
potential. {\em Reasoning by Superposition}
\citep{zhu2025reasoningbysuperposition}  shows that continuous thought
vectors can act as superposition states, encoding multiple search
frontiers in parallel and enabling efficient breadth-first reasoning.
This construction allows a shallow transformer to solve problems such as
directed graph reachability far more efficiently than discrete CoT, which
is forced into sequential exploration and risks being trapped in local
solutions. Complementarily, \emph{Soft Thinking}  \citep{zhang2025softthinkingunlockingreasoning} proposes replacing discrete (``hard'') tokens with concept tokens—probability-weighted mixtures of embeddings—that retain full distributional information. This enables the model to implicitly follow multiple reasoning paths simultaneously, yielding empirical improvements in both accuracy and token efficiency.

Despite these promising claims, the practical benefits of continuous
reasoning at inference time on top of discrete-token base models remain contested. In particular, 
\cite{wu2025llmssinglethreadedreasonersdemystifying} critically re-examine {\em Soft Thinking} and find that vanilla implementations often underperform their discrete counterparts. Their analysis suggests that LLMs, when given soft inputs, default to relying on the single highest-probability token—effectively reducing Soft Thinking to greedy decoding. Further, existing methods for soft thinking are limited to inference on models trained with discrete CoTs.

Training of continuous-token reasoning models has proven to be difficult, either
due to computational constraints from full backpropagation through all
steps of continuous reasoning (this limited the CoT to 6 steps in
\citet{hao2024coconut}), or due to the necessity of strongly grounding
the continuous reasoning into ground-truth discrete reasoning traces
\citep{shen2025codi}.  This is why several of the works above limit
themselves to applying continuous reasoning at inference time without
training \citep{zhang2025softthinkingunlockingreasoning, wu2025llmssinglethreadedreasonersdemystifying}.

\begin{figure}[t]
  \centering
\begin{minipage}[c]{0.45\linewidth}
  \centering
  \scalebox{1}[0.95]{%
    \includegraphics[width=\linewidth]{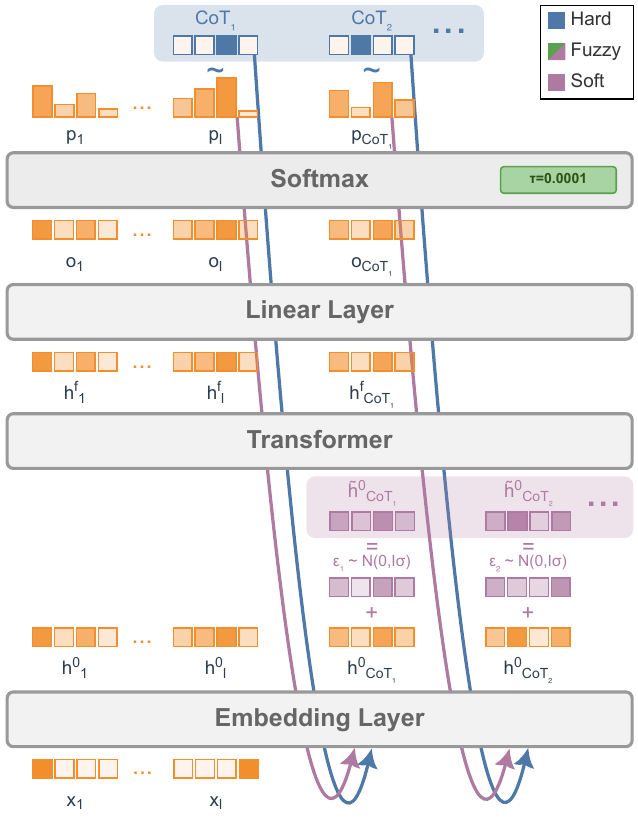}%
  }
\end{minipage}\hfill
  \begin{minipage}[c]{0.45\linewidth}
    \captionsetup{font=small}
    \captionof{figure}{\textbf{Hard, fuzzy and soft generation during CoT phase}. In \emph{hard} generation, at each time step, a discrete token $CoT_t$ is sampled from the probability vector $p_{t-1}$ and its embedding $h^0_{CoT_1}$ is passed to the transformer, generating a sequence of discrete CoT tokens: $CoT_1,..., CoT_T$ over time. In \emph{fuzzy} and \emph{soft} generation, at each time step, noise, $\epsilon_t$, is injected into the probability weighted mixture embedding, $h_0^t=p_{t-1}E$, where $E$ is the token embedding matrix. This noisy input embedding is passed to the transformer, generating a sequence of continuous noisy CoT embeddings: ${\tilde h^0_{CoT_1},..., \tilde h^0_{CoT_T}}$ over time. Additionally, for \emph{fuzzy} generation, the temperature $\tau$ used in the CoT phase tends to 0, such that the non-noisy embeddings $h^0$ reduce to embeddings of discrete tokens. We find that the combination of soft/fuzzy training and hard inference performs universally best, matching hard training at pass@$1$ and surpassing it at pass@$32$, indicating better preservation of diversity.}
    \label{fig:viz}
  \end{minipage}
\confonly{\vspace{-20pt}}
\end{figure}
In this work, we address these limitations by developing an approach to reinforce continuous CoTs with controlled noise, making them amenable to reinforcement learning (RL) training. We theoretically outline two types of continuous CoT learning with {\em soft} and {\em fuzzy} tokens (see Figure \ref{fig:viz}) and provide extensive empirical evidence with Llama-3.x and Qwen-2.5 models trained on a number of mathematical datasets (GSM8K, MATH, DeepScaleR) and evaluate on a variety of mathematical and out-of-domain benchmarks. Our contributions and findings are as follows:
\begin{itemize}
    \item \textbf{A continuous-token post-training algorithm.} We propose
    the first continuous CoT finetuning algorithm that does not require
    ground-truth CoT annotations, directly tackling the challenge of
    learning continuous reasoning representations, at a negligible
    computational overhead compared to discrete CoTs.
    \item \textbf{Pass@1 parity.} We show that continuous CoT post-training is
    competitive with traditional discrete-token CoTs for pass@1 criteria, with a statistically similar performance on most model-dataset combinations. 
    \item \textbf{Pass@32 gains.} 
    Under sampling (pass@32), continuous CoT training outperforms discrete CoT on average, demonstrating greater CoT diversity. 
  \item \textbf{Improved robustness.} Continuous CoT training does
  not degrade the base model's log-likelihood on HellaSwag, ARC and MMLU,
  whereas discrete CoT training does, on average. Further, continuous CoT
  training is robust to collapse observed with discrete training
  on Llama-8B-Instruct with respect to in-distribution and out-of-distribution performance.
   \item \textbf{Hard inference on soft models.} In our experiments, adding continuous CoT \emph{at inference} on top of a hard-token-finetuned model does not bring benefits, contrasting with \citet{zhang2025softthinkingunlockingreasoning}. On the opposite, the best performance is obtained when using discrete CoT at inference on top of a continuous CoT trained model. This means that a practitioner can deploy standard inference methods directly to reap the advantages from models trained with continuous tokens.  
    \item \textbf{Entropy analysis.} We present a detailed analysis of
    the entropy profiles of models, showing how continuous CoT
    fine-tuning more closely preserves the entropy profiles of the base Llama models, compared to discrete CoT.
\end{itemize}

\section{Related Work}

It is natural to question whether token space is the ideal medium for
reasoning, particularly in tasks that demand higher levels of semantic
abstraction. However, unrolling beyond token space during pretraining
departs from the original training distribution and may require
additional mechanisms to help a hard-token–trained LLM adapt to
alternative, potentially more expressive, representations of chains of
thought. Several prior works have confronted this challenge.
\citep{goyal2024pausetokens} proposes inserting dedicated placeholder
tokens, ``Pause Tokens", into rollouts to encourage more deliberate
“thinking” in the internal layers. 
 Coconut \citep{hao2024coconut}, in contrast, explicitly ventures
into continuous-space reasoning by distilling ground-truth chains of thought
into continuous tokens, but its benefits appear confined to a
benchmark designed for this purpose. Furthermore, Coconut requires
ground truth chains-of-thought and a gentle ``hard-to-soft" distillation
schedule, and comes with computational constraints that have limited it
to 6 continuous CoT tokens.

A number of follow-up works have tried to tackle these challenges (see \citet{zhu2025surveylatentreasoning} for a survey on latent-space reasoning). 
Given the frequent difficulty of making an LLM accept continuous tokens when they are not native to their training, it makes sense to
distinguish {\em when} in the pipeline continuous tokens are invoked (at pretraining, at post-training or at inference). We list those works closest to ours, while necessarily being incomplete as the scope widens; we recommend the survey by \citet{zhu2025surveylatentreasoning} for a taxonomy.

\paragraph{\textbf{Inference.}} \citet{zhang2025softthinkingunlockingreasoning}
introduce so-called {\em soft tokens} as the softmax layers of the
output tokens, and propose to perform inference in soft-token space.
Combined with a few other interventions (a hand-crafted stopping
criterion, and only keeping the few top dimensions of embedding space)
they seem to gain some performance on some benchmarks purely through
inference-time interventions. A more measured view of the same is given
in \citet{wu2025llmssinglethreadedreasonersdemystifying}, where the
previous results are not confirmed, unless \emph{noise} is introduced at
test time into the soft token generation. Noise is important in our
approach for a different reason, to provide exploration for reinforcement
learning.

\paragraph{\textbf{Post-training.}} A number of works aim to introduce continuous tokens during post-training as already mentioned for Coconut \citep{hao2024coconut}.
Codi \citep{shen2025codi} distill a standard discrete CoT model into a
continuous CoT model, by keeping both the emitted tokens and the internal
activities of the continuous model close to that of the original model.

\paragraph{\textbf{Pretraining.}} Several works propose changes to pretraining to incorporate some notion of ``thinking tokens" or latent tokens, and attempt to replace the CoT by some internal activations
inside, or added to, the transformer.  ``Filler tokens'' approaches \citep{lanham2023measuringfaithfulnesschainofthoughtreasoning,goyal2024pausetokens,pfau2024fillertokensdotbydot,ringel2025learningcontinuethinkingtokenenhanced}
 introduce some bland tokens so that the model can use its
continuous internal activations to reason while reading the bland tokens.
Going further, 
CoCoMix \citep{tack2025cocomixllmpretrainingcontinuousconcepts} intersperses continuous tokens with hard tokens at pretraining, and uses a pretrained sparse autoencoder to couple the hard and soft tokens. 
``Looped transformers'' \citep{saunshi2025loopedtransformers} and ``recurrent depth'' \citep{geiping2025scalingtesttimecomputelatent}
deploy internal, continuous CoTs in the depth direction of the
transformer, by repeating some internal blocks and making the CoT depth
potentially infinite before every token. 

\paragraph{\textbf{Theoretical arguments.}} \citet{zhu2025reasoningbysuperposition}
make a strong argument that continuous CoTs are more expressive than
discrete CoTs, on a natural toy problem (reachability in graphs). They
prove that a 2-layer transformer with continuous CoT can solve the
problem in $O(n)$ vs $O(n^2)$ with discrete tokens. Continuous CoTs are
provably able to use superpositions that explore several reasoning paths
at the same time.  The experiments align with the theoretical
predictions: the superposition is learned in practice.


\section{Method}
\label{sec: method}

\paragraph{\textbf{In a nutshell.}} Similarly to Coconut or Soft Tokens,
our method keeps the full probability distribution after the softmax
step, instead of emitting a discrete token. This mixture of tokens is fed
to the input of the transformer for the next step. Contrary to prior
work, we then inject noise on the input embeddings. This noise produces
the necessary exploration to apply RL fine-tuning. In contrast, Soft
Tokens do not fine-tune, and Coconut relies on backpropagation through
time through the whole continuous CoT (coming with strong computational
limitations) plus a curriculum to ground the continuous CoTs into
ground truth discrete CoTs.
We now describe our method in detail.

\paragraph{\textbf{Notation for transformers.}}
We decompose a standard LLM architecture into the following blocks. 

We denote by $V$ the vocabulary size. Thus, each token can be seen as a
one-hot vector $x_t\in \R^V$, and the sequence of tokens $x_{<t}$ up to
time $t$ can be seen as a matrix of size $t\times V$.

We denote by $E$ the token embedding matrix of an LLM, that takes a
sequence of tokens $x_{<t}$ and returns a sequence of embeddings
$h^0_{<t}$ by a linear, token-wise mapping
\begin{equation}
h^0_{<t}=x_{<t}E.
\end{equation}
Denoting by $n_0$ the dimension of the input
embedding of the transformer stack, $E$ is a matrix of size $V\times
n_0$.

We denote by $\transf$ the
transformer stack, that turns the sequence of input embeddings $h^0_{<t}$
into a sequence of output embeddings
\begin{equation}
h^L_{<t}=\transf(h^0_{<t})
\end{equation}
where $L$ is the depth of the transformer stack. $h^L_{<t}$ is a matrix
of size $t\times n_L$ with $n_L$ the dimension of the output
layer of the transformer stack.

Probabilities for next-token prediction are obtained as follows. The
output encodings $h^L_{<t}$ are turned into logits by a decoding matrix
$W_d$ of size $n_L\times V$ where $n_L$ is the output dimension of the
transformer stack. The next-token probabilities are obtained by applying
a softmax at temperature $\tau\geq 0$ to these logits:
\begin{equation}
p_{<t}=\softmax(h^L_{<t}\, W_d/\tau)
\end{equation}
where $p_{<t}$ is a matrix of size $t\times V$, 
the softmax is applied for each $t$ independently, and softmax is
extended by continuity for temperature $\tau=0$.

\paragraph{\textbf{Standard (hard) tokens.}}
In standard, hard-token models, each
token $x_t$ is a one-hot vector $x_t\in \R^V$. At inference time,
to compute a next-token prediction $x_t$ given the sequence of previous
tokens $x_{<t}$, one first computes the next-token probabilities $p_{<t}$
given $x_{<t}$ as above. Then the next token is sampled according to the
last component $p_{t-1}$ of $p_{<t}$:
\begin{equation}
\Pr(x_t=1_i)=p_{t-1,i}
\end{equation}
where $1_i$ denotes the one-hot encoding of token $i$. This is applied
inductively to get the sequence of next tokens.

\paragraph{\textbf{Soft thinking.}} In soft thinking \citep{zhang2025softthinkingunlockingreasoning,wu2025llmssinglethreadedreasonersdemystifying}, during the CoT phase, instead of sampling a next token $x_t$ according to the
probabilities $p_t$, the probabilities are directly used to define a
mixture of embeddings. The next input layer embedding is obtained as
\begin{equation}
h^0_t = \sum_i \Pr(x_t=1_i)e_i= p_{t-1}E
\end{equation}
where $e_i$ is the embedding for token $i$. Then the transformer stack is
applied normally to $h^0$. \footnote{The model used in Coconut
\citep{hao2024coconut} is slightly different in that it directly feeds the
output embedding as next-step input embeddings, namely, $h^0_t=
h^L_{t-1}$, assuming dimensions are the same. This bypasses the expansion
from hidden dimension to vocabulary size and back, as well as the softmax
and temperature.}

After the CoT phase is done, the model samples normal (hard) tokens.

This model is not amenable to direct RL training via Reinforce-like
algorithms, because of the absence of noise or random choices: the whole
CoT is a deterministic and differentiable function of the prompt. In
principle, it could be optimized directly by backpropagating through all
the timesteps of the CoT, similarly to Backpropagation Through Time
(BPTT). But this leads to technical and memory challenges that we will
not discuss here. 

\paragraph{\textbf{Noisy soft thinking: soft tokens and fuzzy tokens.}} Instead,
we propose to make soft thinking trainable by RL, just by introducing
noise into the soft thinking process. We simply add noise to the
computation of $h^0_t$:
\begin{equation}
\tilde h^0_t= p_{t-1}E+\sigma N(0,\Id)
\end{equation}
with some standard deviation $\sigma>0$.
Then, at the next timestep, the transformer stack is fed $\tilde h^0_t$.

We also experimented with adding noise at other places, such as on the
logits (Appendix~\ref{app:ablations:noise-placement}).

We call this model \emph{soft tokens}; we use the term \emph{fuzzy
tokens} when the temperature $\tau$ used
during the chain of thought tends to $0$, because in that case, the
non-noisy embeddings $h^0$ reduce to embeddings of true discrete tokens,
so $\tilde h^0$ are normal discrete tokens embeddings up to
noise $\sigma$.

\paragraph{\textbf{Reinforcement learning on soft tokens.}} Introducing
exploration noise on the soft CoT tokens makes it possible to optimize the
model via reinforcement learning. (For traditional discrete CoT tokens,
exploration comes from the random sampling of a token from the softmax
probabilities.)

We describe here the derivation of Reinforce for noisy
soft thinking. More advanced Reinforce-like methods such as RLOO, GRPO,
PPO... are derived from Reinforce in the standard way.

Given a prompt, we sample a soft CoT, then sample a final answer $a$
given the CoT. Let $R(a)$ denote the reward obtained for an answer $a$.
The objective is to maximize the expected reward.

The CoT sampling is fully defined by the sampling of the noisy soft
tokens $\tilde h^0$. Therefore, the objective is to maximize the
expectation
\begin{equation}
\E_{(\tilde h,a)\sim \pi}[R(a)]
\end{equation}
for a given prompt,
where $\pi$ is the current model.

By the standard Reinforce theorem \cite{sutton1998}, this is equivalent to minimizing the
loss
\begin{equation}
\E_{(\tilde h,a)\sim \pi\sg} \left[-R(a) \left(
\log \pi(\tilde h^0)+\log \pi(a|\tilde h^0)
\right)\right]
\end{equation}

The term $\log \pi(a|\tilde h^0)$ just represents fine-tuning the answer
given the CoT, and can be computed in a standard way, since sampling of
$a$ is done in a standard way.

The term $\log \pi(\tilde h^0)$ can be decomposed as a sum over
timesteps,
\begin{equation}
\log \pi(\tilde h^0)=\sum_t \log \pi(\tilde h^0_t|\tilde h^0_{<t})
\end{equation}
and each of those terms can be computed easily: indeed, knowing the soft
tokens $\tilde h^0_{<t}$, we can compute the non-noisy next-token input
embedding $h^0_t$. Since the noise is Gaussian, we just have:
\begin{equation}
\log \pi(\tilde h^0_t|\tilde h^0_{<t})=-{\textstyle
\frac{1}{2\sigma^2}}\norm{\tilde h^0_t-h^0_t}^2+\mathrm{cst}
\end{equation}
and we note that $h^0_t$ is a differentiable function of the previous
soft tokens $\tilde h^0_{<t}$, depending on the parameters of the model.

This makes it possible to apply the family of Reinforce-like algorithms
to noisy soft tokens.

Computational overhead is minimal: storing the probability vector $p_t$ at each
step (vector of size $V$), and injecting noise on the first layer.

\section{Experiments}
\label{sec:experiments}

\paragraph{\textbf{Models tested.}}
We train three variations of CoT models described in Section~\ref{sec: method}:
\begin{itemize}
    \item Hard tokens: Categorical sampling of ordinary hard CoT tokens
    with temperature $\tau=1.0$.
    \item Soft tokens: Instead of sampling hard tokens, we use the full
    probability mixture at temperature $\tau=0.5$ to compute embeddings, and
    add Gaussian noise to the embeddings.
    \item Fuzzy tokens: Like soft tokens, but at temperature
    $\tau=0.0001$, which brings them very close to hard tokens embeddings, to which
    we add Gaussian noise.
\end{itemize}
For the scale of the Gaussian noise, we set this equal to 0.33 times the
root-mean-square norm of the token embeddings, so that the noise is
comparable but a bit smaller than the embeddings. In practice we observe
our algorithm is robust to ratios less than or equal to $1.0$
(Appendix~\ref{app:ablations:noise-scale}). In further ablations (Appendix~\ref{app:ablations:temperature}), we observe that our algorithm is also robust to temperature values $\tau \in [0.0001,\,0.1]$.

\paragraph{\textbf{Inference settings.}}
At test time, we decouple the inference method from the training method:
for each trained model (hard, soft, fuzzy), 
we evaluate six inference settings.
We vary the decoding of the CoT
as follows,
but with answers always greedily decoded at temperature 0:
\begin{itemize}
    \item Hard Greedy: discrete tokens, CoT temperature $\tau=0.0$ at test time
    \item Hard Sample: discrete tokens, CoT temperature $\tau=1.0$ at test time
    \item Soft Greedy: Gaussian scale $\sigma=0.0$, CoT temperature $\tau=0.5$ at test
    time
    \item Soft Sample: Gaussian scale $\sigma=0.33*\text{root-mean-square norm}$, CoT
    temperature $\tau=0.5$ at test time
    \item Fuzzy Greedy: Gaussian scale $\sigma=0.0$, CoT temperature $\tau=0.0001$ at
    test time
    \item Fuzzy Sample: Gaussian scale $\sigma=0.33*\text{root-mean-square norm}$, CoT
    temperature $\tau=0.0001$ at test time
\end{itemize}
For instance, ``soft'' training with ``hard greedy'' testing amounts to
training with a mixture at $\tau=0.5$ with Gaussian noise, then applying
the model with hard tokens at test time.

The ``sample'' settings are the same as the variants used during
training for the CoT, while the ``greedy'' setting more aggressively
target the mode of the distribution at each step of the CoT.

\paragraph{\textbf{Reinforce with group baseline.}} We fine-tune the models with
RLOO, namely, Reinforce
using a per-prompt leave-one-out (LOO) group baseline \citep{rloo}:
for each sample and reward, we subtract the average reward
obtained on the other samples for the same prompt. We include the RLOO
loss in Appendix~\ref{sec:rloo} for completeness.

At each update we draw a mini-batch of \(B=2\) distinct prompts
\(\{x_b\}_{b=1}^B\). For each prompt we sample \(G=32\) sequences
\(y_{b,g}\) that contain a chain-of-thought (CoT) followed by a final
answer. The prompt instructs the model to end the CoT with ``\verb|The final answer is: |'' (see Appendix~\ref{app:task-prompt}).

Rewards are computed only on the final answer using the \emph{Math Verify} package \citep{mathverify} against the ground-truth label: 
\begin{align}
r_{b,g} \;=\;
\begin{cases}
100, & \text{if }\operatorname{Verify}(a_{b,g})=1,\\[2pt]
10,  & \text{if }\operatorname{Verify}(a_{b,g})=0 \ \text{and}\ \operatorname{ExtractBoxed}(a_{b,g})=1,\\[2pt]
0,   & \text{otherwise.}
\end{cases}
\end{align}

\paragraph{\textbf{Datasets and base models.}}
We train Llama 3.2 3b Instruct, Llama 3.1 8b Instruct
\citep{dubey2024llama} and Qwen 2.5 3b Instruct \citep{Yang2024Qwen25TR}
on math reasoning datasets including GSM8K \citep{cobbe2021gsm8k}, MATH
(\cite{hendrycks2021measuring} and DeepScaleR \citep{deepscaler2025}.

For each model trained on a dataset, we evaluate test performance on three math reasoning test
datasets: GSM8K, MATH and OlympiadBench
\citep{he2024olympiadbenchchallengingbenchmarkpromoting}. For
OlympiadBench, following prior work, we use the 675 subset of math
questions which have final answers and do not contain images or figures. Similarly, for MATH, we evaluate on the MATH-500 \citep{math500_hf} subset of the MATH test set. To assess out-of-distribution generalization, 
we also test the resulting models on standard benchmarks:
HellaSwag \citep{Zellers2019HellaSwagCA}, MMLU \citep{hendryckstest2021mmlu},
and ARC/AI2 Reasoning Challenge \citep{clark2018arc}.

On each training dataset, we train for 4k steps and monitor greedy
validation performance; the final model used for testing is the best
performing under our greedy validation performance. (For hard CoTs, greedy
performance refers to greedily decoding CoTs and answers. For soft and
fuzzy CoTs, greedy performance refers to decoding CoTs with no Gaussian
noise and greedily decoding answers.) During training, we sample a maximum
of 128 CoT tokens for GSM8K and 512 CoT tokens for MATH and DeepScaler
respectively, under our early stopping criterion (see
Appendix~\ref{app:stopping-criterion}); on all datasets we sample 32 answer tokens.
For evaluation, in all cases, we sample a maximum of 512 CoT tokens under our early stopping criterion followed by 32 answer tokens.

Each setup was run with $3$ independent random seeds; the tables
report the resulting mean and standard deviation. Training and validation success rates may be found in Figure \ref{fig:training_llama_3b_gsm8k}. For details on hyper-parameters, see Appendix~\ref{app:hyperparam-search}.

\begin{figure}
    \centering
    \includegraphics[width=0.95\linewidth]{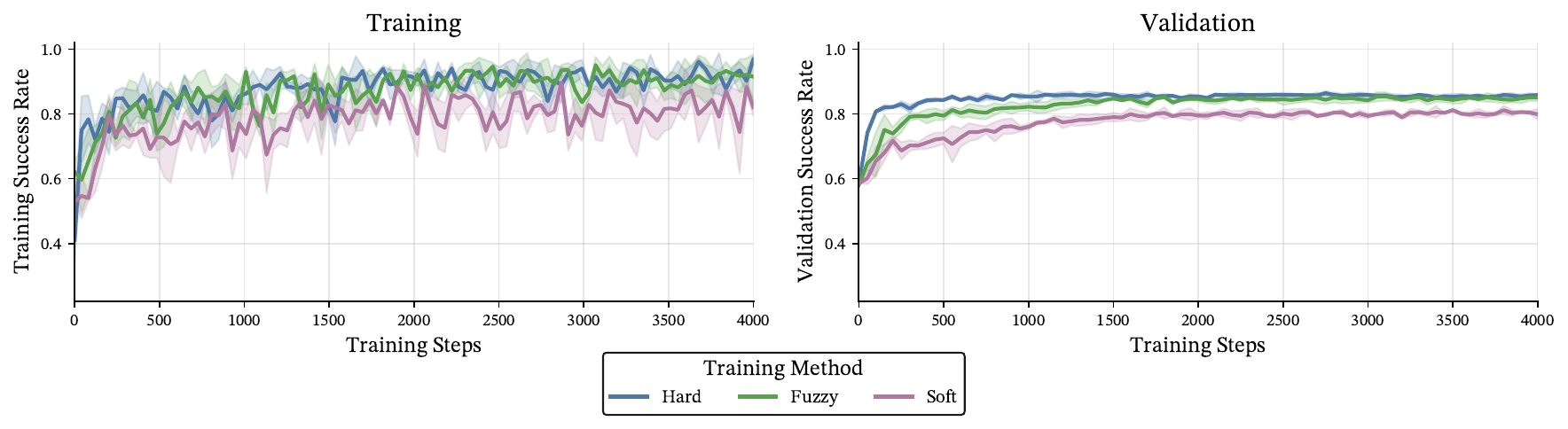}
    \caption{\textbf{Llama 3b Instruct trained on GSM8K} (a) Training performance across steps; one step = two prompts $\times$ 32 samples each. (b) Greedy validation performance used for model selection. For the remaining trained models, see Appendix \ref{app:supplementary-results:training}.}
    \label{fig:training_llama_3b_gsm8k}
\end{figure}

\paragraph{\textbf{Results.}}
\begin{figure}
    \centering
    \includegraphics[width=0.95\linewidth]{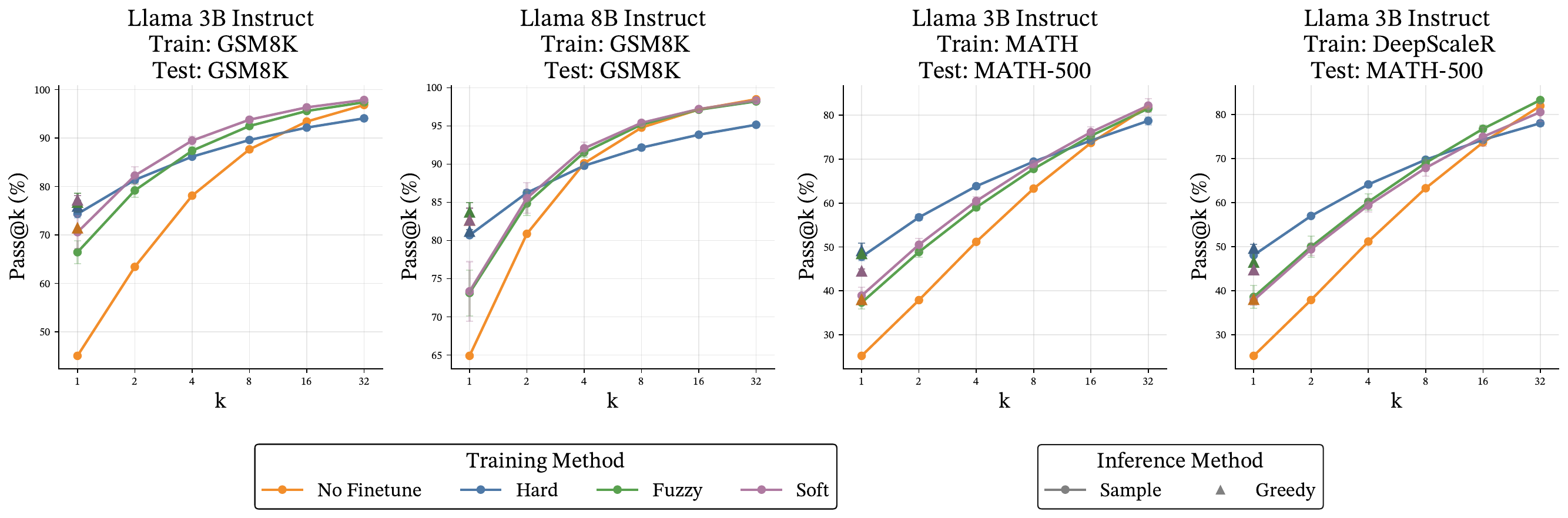}
    \caption{\textbf{Hard inference pass@k for Llama models} (for soft/fuzzy inference and Qwen see Appendix \ref{app:supplementary-results:pass@k}). \emph{We observe soft/fuzzy training improves pass@$32$, pointing to preserved diversity. Greedy Pass@1 (the triangles) for all training methods are clustered together.}}
    \label{fig:pass@k-main}
\end{figure}
Across datasets and models, \emph{the three training schemes are broadly
comparable}, generally achieving similar greedy pass@1 performance as shown in Table \ref{tab:main}.
This demonstrates that fuzzy and soft training are effective. On the
other hand, \emph{soft and fuzzy training have a clear overall advantage for
pass@32} over hard training (this signal is clearest on Llama, as shown in Figure \ref{fig:pass@k-main}).

We observe a gap between the \emph{hard greedy} and \emph{hard sample} inference settings for the base models and models trained with fuzzy/soft CoTs, whereas the gap for models trained with hard CoTs is very small (see Figure \ref{fig:pass@k-main} and Table \ref{tab:main}). We note that, for the base and fuzzy/soft trained models, $\tau=1$ is evidently not optimal for pass$@1$ (greedy $\tau=0$ is always better). At low values of $k$, the optimal $\tau$ may be some interpolation between $0$ and $1$. Further, for pass@32, we also observe a closing of the gap between the trained models and base models. This loss of diversity is well reported for Reinforce style algorithms where the reward is based on answer correctness alone \citep{song2025outcomebasedexplorationllmreasoning}. 

For all training methods (hard, soft or fuzzy), \emph{hard inference
generally performs best}, both for pass@1 and pass@32 (see Appendix \ref{app:soft_and_fuzzy}). In particular, we
do \emph{not confirm previously reported benefits of soft inference on
hard (normal) training} \citep{zhang2025softthinkingunlockingreasoning}.

\begin{savenotes}
\begin{table}[ht]\centering\footnotesize
\resizebox{\textwidth}{!}{%
\begin{tabular}{l l|c c c|c c c|c c c}
\toprule
& & \multicolumn{3}{c|}{GSM8K} & \multicolumn{3}{c|}{MATH-500} & \multicolumn{3}{c}{OlympiadBench} \\
Model & Training &
\makecell{Greedy\\pass@1} &
\makecell{Sample\\pass@1} &
\makecell{Sample\\pass@32} &
\makecell{Greedy\\pass@1} &
\makecell{Sample\\pass@1} &
\makecell{Sample\\pass@32} &
\makecell{Greedy\\pass@1} &
\makecell{Sample\\pass@1} &
\makecell{Sample\\pass@32} \\
\midrule
\rowcolor{paleviolet}
\multirow{10}{*}{\cellcolor{white}\makecell{llama 3b\\instruct}}
    & no finetune & 71.4$\pm$0.0 & 45.0$\pm$0.0 & 96.8$\pm$0.0 & 38.0$\pm$0.0 & 25.2$\pm$0.0 & 82.0$\pm$0.0 & 17.9$\pm$0.0 & 12.0$\pm$0.0 & 52.3$\pm$0.0 \\
\cmidrule(l){2-11}
    & gsm8k hard & \greedy{75.9$\pm$1.3} & 74.3$\pm$0.8 & 94.1$\pm$0.3 & 34.6$\pm$0.2 & 31.3$\pm$0.4 & 72.4$\pm$1.1 & 10.6$\pm$0.9 & 9.4$\pm$0.5 & 39.3$\pm$1.3 \\
    & gsm8k fuzzy & \greedy{76.7$\pm$1.8} & 66.4$\pm$2.4 & \sample{97.4$\pm$0.3} & \greedy{42.6$\pm$3.2} & 32.4$\pm$1.7 & 83.2$\pm$1.1 & 17.4$\pm$0.9 & 13.0$\pm$1.1 & 55.2$\pm$1.2 \\
    & gsm8k soft & \greedy{77.2$\pm$0.9} & 70.6$\pm$3.4 & \sample{97.9$\pm$0.3} & \greedy{43.5$\pm$1.4} & 37.2$\pm$1.0 & \sample{84.8$\pm$0.4} & \greedy{18.9$\pm$0.2} & 15.5$\pm$0.7 & \sample{58.6$\pm$1.6} \\
\cmidrule(l){2-11}
    & math hard & \greedy{80.0$\pm$0.5} & \greedy{79.7$\pm$0.8} & 96.7$\pm$0.1 & \greedy{49.1$\pm$1.7} & \greedy{47.8$\pm$0.9} & 78.7$\pm$0.9 & \greedy{22.7$\pm$1.1} & \greedy{22.2$\pm$0.4} & 49.3$\pm$1.3 \\
    & math fuzzy & \greedy{79.6$\pm$1.4} & 68.5$\pm$2.1 & \sample{97.6$\pm$0.3} & \greedy{48.5$\pm$0.3} & 37.3$\pm$1.5 & \sample{81.5$\pm$0.7} & \greedy{21.6$\pm$0.6} & 15.9$\pm$0.8 & 52.8$\pm$1.6 \\
    & math soft & 76.8$\pm$1.0 & 70.7$\pm$2.4 & \sample{97.8$\pm$0.3} & 44.5$\pm$0.6 & 38.9$\pm$1.9 & \sample{82.2$\pm$1.6} & 19.2$\pm$1.1 & 16.6$\pm$0.4 & \sample{55.8$\pm$1.2} \\
\cmidrule(l){2-11}
    & deepscaler hard & \greedy{79.7$\pm$1.3} & \greedy{79.6$\pm$0.2} & 96.6$\pm$0.3 & \greedy{49.6$\pm$0.9} & 48.1$\pm$0.4 & 78.1$\pm$0.7 & \greedy{23.2$\pm$1.4} & \greedy{23.8$\pm$0.6} & \sample{50.8$\pm$1.5} \\
    & deepscaler fuzzy & \greedy{78.8$\pm$0.8} & 69.8$\pm$4.7 & \sample{97.7$\pm$0.5} & 46.5$\pm$2.2 & 38.6$\pm$2.6 & \sample{83.3$\pm$0.3} & 19.5$\pm$1.0 & 16.6$\pm$1.4 & \sample{54.4$\pm$2.1} \\
    & deepscaler soft & \greedy{77.9$\pm$1.8} & 71.0$\pm$1.5 & \sample{98.0$\pm$0.1} & 44.7$\pm$1.2 & 37.9$\pm$1.3 & 80.6$\pm$2.2 & 21.0$\pm$0.6 & 16.4$\pm$0.7 & \sample{53.3$\pm$1.7} \\
\midrule
\rowcolor{paleviolet}
\multirow{4}{*}{\cellcolor{white}\makecell{llama 8b\\instruct}}
    & no finetune & 82.6$\pm$0.0 & 64.9$\pm$0.0 & 98.5$\pm$0.0 & 44.4$\pm$0.0 & 31.1$\pm$0.0 & 79.8$\pm$0.0 & 19.6$\pm$0.0 & 11.7$\pm$0.0 & 52.6$\pm$0.0 \\
\cmidrule(l){2-11}
    & gsm8k hard & 81.2$\pm$0.2 & 80.7$\pm$0.3 & 95.1$\pm$0.2 & 20.2$\pm$0.8 & 19.8$\pm$1.4 & 45.4$\pm$3.2 & 3.8$\pm$0.5 & 3.4$\pm$0.5 & 16.4$\pm$2.4 \\
    & gsm8k fuzzy & \greedy{83.7$\pm$1.3} & 73.1$\pm$3.0 & \sample{98.2$\pm$0.2} & \greedy{44.6$\pm$2.1} & 33.8$\pm$2.7 & \sample{83.1$\pm$0.9} & \greedy{18.0$\pm$1.4} & 12.5$\pm$1.3 & \sample{56.0$\pm$2.6} \\
    & gsm8k soft & \greedy{82.6$\pm$1.6} & 73.3$\pm$3.9 & \sample{98.3$\pm$0.2} & \greedy{44.7$\pm$2.3} & 34.8$\pm$2.2 & \sample{83.9$\pm$1.1} & \greedy{17.9$\pm$1.0} & 13.1$\pm$0.7 & \sample{56.8$\pm$1.2} \\
\midrule
\rowcolor{paleviolet}
\multirow{4}{*}{\cellcolor{white}\makecell{qwen 3b\\instruct}}
    & no finetune & 8.9\footnotemark
$\pm$0.0 & 17.2$\pm$0.0 & 95.1$\pm$0.0 & 29.0$\pm$0.0 & 25.5$\pm$0.0 & 81.0$\pm$0.0 & 16.9$\pm$0.0 & 14.3$\pm$0.0 & 50.2$\pm$0.0 \\
\cmidrule(l){2-11}
    & math hard & \greedy{84.0$\pm$0.8} & 83.0$\pm$0.2 & 97.2$\pm$0.3 & \greedy{59.0$\pm$1.7} & 57.1$\pm$0.1 & \sample{83.6$\pm$1.0} & \greedy{29.4$\pm$0.3} & 27.8$\pm$0.5 & \sample{61.1$\pm$0.6} \\
    & math fuzzy & \greedy{84.4$\pm$0.8} & 81.6$\pm$0.9 & \sample{98.1$\pm$0.2} & \greedy{58.1$\pm$0.9} & 55.5$\pm$1.2 & \sample{84.4$\pm$0.2} & 27.2$\pm$0.5 & 24.4$\pm$0.7 & \sample{60.7$\pm$0.5} \\
    & math soft & \greedy{82.9$\pm$0.9} & 78.7$\pm$2.6 & \sample{97.6$\pm$0.5} & 54.7$\pm$0.3 & 52.2$\pm$1.1 & \sample{84.4$\pm$0.7} & 24.3$\pm$1.8 & 22.0$\pm$0.5 & 58.5$\pm$1.0 \\
\bottomrule
\end{tabular}%
}
\caption{\textbf{Results of hard inference on GSM8K, MATH-500 and OlympiadBench test sets}. In \greedy{blue} the best pass@$1$ performance and in \sample{green} the best pass@$32$ for each (base model, training set) pair. \emph{We observe broadly comparable performance at pass@$1$ and improved soft/fuzzy training pass@$32$}. For comparisons of hard, fuzzy and soft inference, see Tables \ref{tab:gsm8k}, \ref{tab:math500}, \ref{tab:olympiad} in Appendix \ref{app:soft_and_fuzzy}.}
\label{tab:main}
\end{table}
\end{savenotes}

One setup stands out: \emph{when training Llama-8B-Instruct on GSM8K
and testing on MATH-500, only fuzzy and soft-trained models achieve good
scores}, while classical hard token fine-tuning is ineffective.
Namely, gsm8k-hard training sharply underperforms on out-of-distribution MATH
(hard-greedy at 20.2\% and pass@$32$ at 45.4\%), whereas gsm8k-fuzzy and gsm8k-soft
trainings recover to 44.6–44.7\% greedy and 83.1–83.9\% pass@32 while maintaining in-distribution performance on GSM8K.
Llama-8B-Instruct has a good performance from the start on GSM8K
(presumably because it was exposed to this dataset), but
this does not translate to good performance on MATH. Further hard
fine-tuning makes things worse, but further soft or fuzzy
fine-tuning on GSM8K does bring improvement on MATH.
Thus, \emph{fuzzy and soft training appear to bring more generalization on
Llama-8B-Instruct.}

\paragraph{\textbf{Out-of-domain robustness.}} One risk of LLM fine-tuning on a
dataset is degrading the general performance of the model on other
datasets. To assess this, we test the trained models on three standard
benchmarks in Table~\ref{table:stdbenchmarks}. We report both the success
rate (with hard greedy sampling) and the negative log-likelihood per token (NLL) of the correct answer.

\begin{table}[ht]\centering\footnotesize
    \resizebox{0.85\textwidth}{!}{%
\begin{tabular}{l l|c c|c c|c c}
\toprule
& & \multicolumn{2}{c|}{Hellaswag} & \multicolumn{2}{c|}{ARC} & \multicolumn{2}{c}{MMLU} \\
Model & Training & \makecell{Accuracy} & \makecell{NLL Correct} &
 \makecell{Accuracy} &  \makecell{NLL Correct} &
 \makecell{Accuracy} & \makecell{NLL Correct} \\
\midrule
\rowcolor{paleviolet}
\multirow{10}{*}{\cellcolor{white}\makecell{llama 3b\\instruct}}
    & no finetune & 66.46$\pm$0.00 & 2.53$\pm$0.00 & 72.79$\pm$0.00 & 2.86$\pm$0.00 & 60.85$\pm$0.00 & 1.61$\pm$0.00 \\
\cmidrule(l){2-8}
 & math hard & 66.20$\pm$0.10 & 2.57$\pm$0.00 & \greedy{73.45$\pm$0.43} & 3.19$\pm$0.03 & 61.11$\pm$0.11 & 1.65$\pm$0.01 \\
 & math fuzzy & \greedy{67.00$\pm$0.12} & \sample{2.53$\pm$0.00} & \greedy{73.25$\pm$0.32} & \sample{2.88$\pm$0.03} & 61.16$\pm$0.04 & \sample{1.60$\pm$0.02} \\
 & math soft & \greedy{67.12$\pm$0.06} & \sample{2.53$\pm$0.01} & 72.76$\pm$0.18 & \sample{2.92$\pm$0.02} & \greedy{61.20$\pm$0.14} & \sample{1.59$\pm$0.02} \\
\cmidrule(l){2-8}
 & deepscaler hard & 66.18$\pm$0.26 & 2.58$\pm$0.01 & \greedy{73.99$\pm$0.57} & 3.18$\pm$0.03 & 60.66$\pm$0.37 & 1.66$\pm$0.01 \\
 & deepscaler fuzzy & \greedy{66.88$\pm$0.06} & 2.55$\pm$0.01 & \greedy{73.05$\pm$0.62} & 2.94$\pm$0.03 & \greedy{61.20$\pm$0.20} & \sample{1.60$\pm$0.02} \\
 & deepscaler soft & \greedy{66.48$\pm$0.46} & \sample{2.56$\pm$0.04} & 72.85$\pm$0.39 & \sample{2.93$\pm$0.08} & \greedy{60.34$\pm$1.20} & \sample{1.68$\pm$0.12} \\
\cmidrule(l){2-8}
 & gsm8k hard & 66.44$\pm$0.18 & 2.57$\pm$0.00 & \greedy{73.53$\pm$0.25} & 3.16$\pm$0.01 & \greedy{61.48$\pm$0.20} & 1.65$\pm$0.01 \\
 & gsm8k fuzzy & \greedy{66.97$\pm$0.21} & \sample{2.53$\pm$0.01} & 73.10$\pm$0.15 & \sample{2.89$\pm$0.01} & \greedy{61.30$\pm$0.21} & \sample{1.60$\pm$0.02} \\
 & gsm8k soft & \greedy{67.13$\pm$0.19} & \sample{2.53$\pm$0.01} & 72.99$\pm$0.36 & \sample{2.89$\pm$0.04} & \greedy{61.36$\pm$0.30} & \sample{1.59$\pm$0.01} \\
\midrule
\rowcolor{paleviolet}
\multirow{4}{*}{\cellcolor{white}\makecell{llama 8b\\instruct}}
    & no finetune & 74.24$\pm$0.00 & 2.34$\pm$0.00 & 81.20$\pm$0.00 & 2.82$\pm$0.00 & 69.00$\pm$0.00 & 1.40$\pm$0.00 \\
\cmidrule(l){2-8}
 & gsm8k hard & \greedy{74.44$\pm$0.10} & 2.37$\pm$0.00 & \greedy{81.49$\pm$0.15} & 3.13$\pm$0.03 & \greedy{68.90$\pm$0.04} & 1.41$\pm$0.00 \\
 & gsm8k fuzzy & \greedy{74.41$\pm$0.03} & \sample{2.35$\pm$0.00} & \greedy{81.63$\pm$0.39} & \sample{2.81$\pm$0.01} & \greedy{68.90$\pm$0.11} & \sample{1.40$\pm$0.00} \\
 & gsm8k soft & \greedy{74.32$\pm$0.08} & \sample{2.35$\pm$0.00} & \greedy{81.46$\pm$0.14} & \sample{2.82$\pm$0.00} & \greedy{68.78$\pm$0.14} & \sample{1.40$\pm$0.01} \\
\midrule
\rowcolor{paleviolet}
\multirow{4}{*}{\cellcolor{white}\makecell{qwen 3b\\instruct}}
    & no finetune & 74.94$\pm$0.00 & 2.29$\pm$0.00 & 83.00$\pm$0.00 & 4.67$\pm$0.00 & 68.08$\pm$0.00 & 1.19$\pm$0.00 \\
\cmidrule(l){2-8}
 & math hard & 75.01$\pm$0.10 & 2.30$\pm$0.00 & \greedy{82.86$\pm$0.40} & 5.62$\pm$0.12 & 67.86$\pm$0.06 & 1.51$\pm$0.02 \\
 & math fuzzy & 75.07$\pm$0.02 & \sample{2.29$\pm$0.00} & \greedy{83.00$\pm$0.28} & \sample{4.74$\pm$0.11} & \greedy{68.16$\pm$0.07} & \sample{1.21$\pm$0.02} \\
 & math soft & \greedy{75.25$\pm$0.12} & \sample{2.30$\pm$0.01} & \greedy{83.12$\pm$0.15} & \sample{5.49$\pm$0.64} & 68.01$\pm$0.06 & 1.33$\pm$0.06 \\
\bottomrule\end{tabular}
    }
    \caption{\textbf{Out-of-domain results}. \emph{Outside of mathematical reasoning domains, fuzzy and soft training on average result in lower negative log-likelihood of the correct answer compared to hard training, indicating a softer touch on the base model's capabilities.}}\label{table:stdbenchmarks}
    \end{table}

The results show comparable success rate for the three training methods
(hard, fuzzy, soft). However, the NLL of the correct
answer is visibly better for fuzzy and soft than for hard, especially on
ARC but also with Qwen on MMLU: \emph{hard training
degrades the base model NLL on out-of-domain datasets, while fuzzy and
soft training preserve it.}

\footnotetext{On Qwen no fine-tune, we observe much lower performance compared to what is reported in the Qwen report \citet{Yang2024Qwen25TR}. The discrepancy is due to our (zero shot) prompting and the resulting CoT generation format, and does not affect our RL trained models (see Appendix~\ref{app:format_following} for details).}

\paragraph{\textbf{Entropy behavior.}}
Next, we report the entropy of the distribution of next-token predictions
during the CoT, as a function of the index within the CoT.

The base Llama models exhibit a very different entropy profile whether
greedy or temperature sampling is used. This shows a difference in
next-token prediction behavior on prefixes sampled from temperature $T=0$
or $T=1$: with the latter, entropy blows up as the CoT progresses with hard sampling, indicating very high
uncertainty as the CoT goes on. Interestingly, \emph{soft or fuzzy sampling on
the base model does not show any entropy blowup}.
The entropy blowup is also not present on Qwen, as seen in Appendix \ref{app:supplementary-results:entropy} Figure \ref{fig:qwen_3b_math_entropy}.
Analyzing base models
is not our main topic, but we still report how this entropy profile
changes after different types of training.

\begin{figure}
    \centering
    \includegraphics[width=0.95\linewidth]{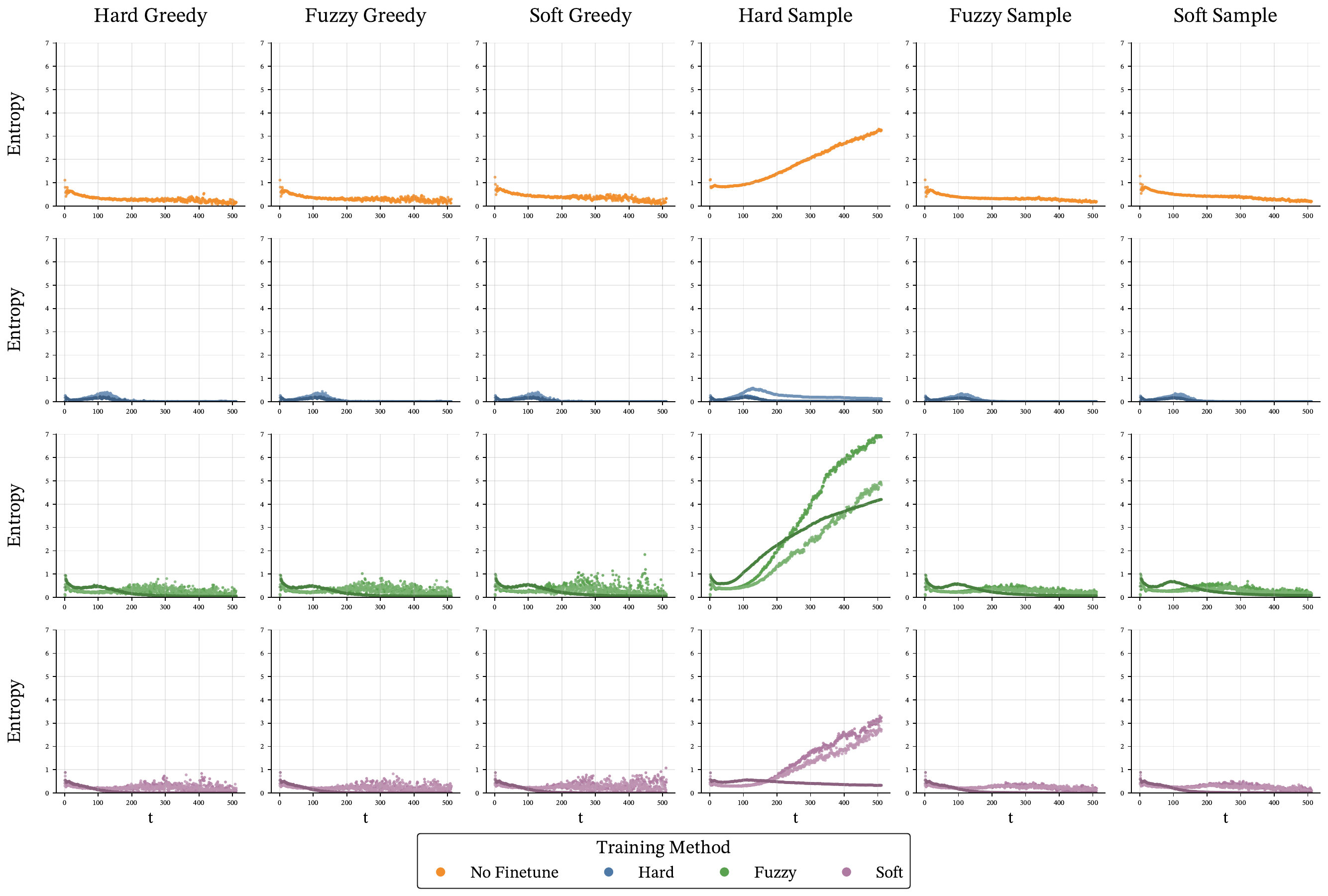}
    \caption{\textbf{Llama 3b Instruct CoT entropy on GSM8K test set}. \emph{Fuzzy and soft training preserves entropy profile of base models; we observe a large change in hard sample profile with hard training.}}
    \label{fig:llama_3b_gsm8k_entropy}
\end{figure}

As observed in Figure \ref{fig:llama_3b_gsm8k_entropy} (and Figures  \ref{fig:llama_3b_deepscaler_entropy}, \ref{fig:llama_8b_gsm8k_entropy},\ref{fig:llama_3b_math_entropy} in Appendix \ref{app:supplementary-results:entropy}), \emph{soft or fuzzy training keep roughly the same entropy profile as the base model}, whether inference is greedy or sampled. On the other hand, hard training
changes the hard sampling entropy profile to resemble greedy sampling on the base model: entropy values are substantially lower. We pose that an explanation for this is that hard training makes the model overconfident, consistent with lower pass@$32$, the worse NLL values we observe on out-of-domain tasks (see Table \ref{table:stdbenchmarks}) and the occasional performance collapse we see for hard training (see Table \ref{tab:main}).

\section{Conclusion}

We have introduced the first reinforcement learning framework for
training continuous Chains-of-Thought in LLMs with minimal computational overhead
and without relying on ground-truth
discrete CoTs.
Across mathematical reasoning benchmarks,
our approach performs on par with 
discrete token training for pass@$1$ success rate and improves pass@$32$
scores. Moreover, it seems to fine-tune the base model with a softer touch, 
better preserving the model's out-of-distribution behavior. This suggests
distinct behavioral differences between soft and hard reasoning
processes.
These results provide evidence that continuous reasoning is not just a theoretical curiosity but a practical alternative for fine-tuning large models.

\paragraph{\textbf{Acknowledgments.}}
We thank Shobhita Sundaram for the helpful discussions. JK acknowledges support by the Simons Foundation through the Collaborative Grant ``The Physics of Learning and Neural Computation''.

\newpage
\bibliography{refs}
\bibliographystyle{plainnat}

\appendix
\newpage
\appendix

\section{Reinforce with Leave-One-Out (RLOO) Baseline}
\label{sec:rloo}

For completeness, we describe here the implementation of RLOO that we
use, directly based on \citep{rloo}.

We fine-tune with Reinforce using a per-prompt leave-one-out (LOO) group
baseline. At each update we draw a mini-batch of \(B\) distinct prompts
\(\{x_b\}_{b=1}^B\). For each prompt we sample \(G\) sequences
\(y_{b,g}\) that contain a chain-of-thought (CoT) followed by a final
answer. Each such sequence provides a reward $r_{b,g}$ as described in
Section~\ref{sec:experiments}.

For each prompt \(x_b\), the LOO group baseline for sample \(g\) averages the other \(G\!-\!1\) rewards from the \emph{same} prompt:
\begin{align}
\bar r_b^{(-g)} \;=\; \frac{1}{G-1}\sum_{\substack{j=1\\ j\neq g}}^{G} r_{b,j},
\qquad
A_{b,g} \;=\; r_{b,g} - \bar r_b^{(-g)} .
\end{align}
Baselines and rewards are treated as constants w.r.t.\ \(\theta\); i.e., stop-gradient.

Let \(y_{b,g,1:T_{b,g}}\) denote all tokens in \(y_{b,g}\) (CoT + answer). With policy \(\pi_\theta\), the per-sequence log-probability is
\begin{align}
\ell_{b,g} \;=\; \sum_{t=1}^{T_{b,g}} \log \pi_\theta\!\big(y_{b,g,t}\mid y_{b,g,<t}, x_b\big).
\end{align}
Our loss is the advantage-weighted negative log-likelihood averaged over batch and group:
\begin{align}
\mathcal{L}(\theta)
&=\;
-\frac{1}{BG}\sum_{b=1}^{B}\sum_{g=1}^{G} A_{b,g}^\mathrm{sg}\,\ell_{b,g},\\
\qquad
\nabla_\theta \mathcal{L}(\theta)
&=\;
-\frac{1}{BG}\sum_{b=1}^{B}\sum_{g=1}^{G} A_{b,g}\!\sum_{t=1}^{T_{b,g}}
\nabla_\theta \log \pi_\theta\!\big(y_{b,g,t}\mid y_{b,g,<t}, x_b\big).
\end{align}
where ${}^{\mathrm{sg}}$ denotes a stop-grad operator on the advantages.

\section{Task Prompt}\label{app:task-prompt}
To guide the model’s behavior during our experiments, we provided the following explicit instruction.  
This prompt ensures that the assistant follows a structured reasoning process before giving the final answer.

\begin{tcolorbox}[colback=gray!10,colframe=black!60,title=Task Prompt,fonttitle=\bfseries]
\texttt{A conversation between User and Assistant. The user asks a question, and the Assistant solves it.  
The assistant first shows the complete reasoning process step by step, then provides the final answer in \textbackslash boxed\{\}.  
The assistant must always follow the format: `User: [question] Assistant: [detailed reasoning] The final answer is: \textbackslash boxed\{[answer]\}.'  
User: QUESTION Assistant: }
\end{tcolorbox}

\section{Stopping Criterion and Prefilling}\label{app:stopping-criterion}

\paragraph{\textbf{Stopping criterion.}}
\begin{itemize}
    \item \emph{Hard:} Monitor the generated text and stop as soon as it ends with
    \texttt{The final answer is: }.
    \item \emph{Soft/Fuzzy:} During continuous generation, form a greedy shadow sequence by taking
    the highest-probability hard token at each step and stop when this shadow ends with
    \texttt{The final answer is: }.
\end{itemize}
If neither condition is met, decoding continues until the maximum chain-of-thought
length \(L\).

\paragraph{\textbf{Prefilling.}}
\begin{itemize}
    \item If early stopping is reached, prefill \texttt{ \textbackslash boxed\{}.
    \item If \(L\) is reached, prefill
    \texttt{The final answer is: \textbackslash boxed\{}.
\end{itemize}

\section{Hyperparameter Search}\label{app:hyperparam-search}
All models were trained with the AdamW optimizer and a cosine learning‐rate schedule using 20 warm up steps.

We tuned all hyperparameters using greedy validation performance. The learning rate and scale factors were used for fuzzy and soft, thus we only sweep over values for fuzzy.

\paragraph{\textbf{Learning rate.}}
For both the hard and fuzzy, we swept over
\(\{1\mathrm{e}{-5},\,9\mathrm{e}{-6},\,\dots,\,2\mathrm{e}{-6},\,1\mathrm{e}{-6}\}\)
for each combination of models.
The same optimal rates were found for both hard and fuzzy:
\begin{itemize}
    \item Llama 3B Instruct: \(6\mathrm{e}{-6}\)
    \item Llama 8B Instruct: \(3\mathrm{e}{-6}\)
    \item Qwen 3B Instruct: \(8\mathrm{e}{-6}\)
\end{itemize}

\paragraph{\textbf{Scale factor (fuzzy only).}}
We additionally swept scale factors
\(\{0.1,\dots,10\}\) on the root mean square embedding norm
across various models and found \(0.33\) to be best, though most values below \(1\) performed well.
An ablation in Section~\ref{app:ablations:noise-scale} supports the finding that our algorithm is robust to scale values below \(1\).

\section{Results on Soft and Fuzzy Inference}\label{app:soft_and_fuzzy}

We evaluate base models and all RL trained models (hard, fuzzy and soft) under six inference settings: hard greedy, hard sample, fuzzy greedy, fuzzy sample, soft greedy and soft sample, described in Section \ref{sec:experiments}. Here, we report test performance on GSM8K (Table \ref{tab:gsm8k}), MATH-500 (Table \ref{tab:math500}) and OlympiadBench (Table \ref{tab:olympiad}). Contrary to previously reported benefits of soft inference on hard trained models \citep{zhang2025softthinkingunlockingreasoning}, we do not observe any improvement in soft inference. Interestingly, we do not observe any gains of soft inference on soft trained models either. In our experiments, hard inference on all models achieves the best performance.

\ifconference
\begin{table}
\centering\footnotesize
\resizebox{\textwidth}{!}{%
\begin{tabular}{l l|c c c|c c c|c c c}
\toprule
& & \multicolumn{9}{c}{Inference Settings}\\
Model & Training &
\makecell{Hard\\Greedy\\pass@1} &
\makecell{Hard\\Sample\\pass@1} &
\makecell{Hard\\Sample\\pass@32} &
\makecell{Fuzzy\\Greedy\\pass@1} &
\makecell{Fuzzy\\Sample\\pass@1} &
\makecell{Fuzzy\\Sample\\pass@32} &
\makecell{Soft\\Greedy\\pass@1} &
\makecell{Soft\\Sample\\pass@1} &
\makecell{Soft\\Sample\\pass@32} \\
\midrule
\rowcolor{paleviolet}
\multirow{10}{*}{\cellcolor{white}\makecell{llama 3b\\instruct}}
    & no finetune & 71.4$\pm$0.0 & 45.0$\pm$0.0 & 96.8$\pm$0.0 & 70.5$\pm$0.0 & 69.3$\pm$0.0 & 93.9$\pm$0.0 & 68.4$\pm$0.0 & 65.2$\pm$0.0 & 94.9$\pm$0.0 \\
\cmidrule(l){2-11}
    & gsm8k hard & \greedy{75.9$\pm$1.3} & 74.3$\pm$0.8 & 94.1$\pm$0.3 & 75.5$\pm$0.6 & 74.7$\pm$0.5 & 92.3$\pm$0.4 & \greedy{75.7$\pm$0.5} & 74.2$\pm$0.4 & 92.6$\pm$0.1 \\
    & gsm8k fuzzy & \greedy{76.7$\pm$1.8} & 66.4$\pm$2.4 & \sample{97.4$\pm$0.3} & \greedy{76.4$\pm$2.1} & \greedy{75.2$\pm$1.8} & 92.0$\pm$1.1 & \greedy{75.1$\pm$1.8} & 73.5$\pm$1.7 & 93.5$\pm$0.8 \\
    & gsm8k soft & \greedy{77.2$\pm$0.9} & 70.5$\pm$3.3 & \sample{97.9$\pm$0.3} & \greedy{76.8$\pm$0.8} & \greedy{76.0$\pm$1.4} & 93.4$\pm$0.3 & 74.5$\pm$1.5 & 74.1$\pm$2.0 & 94.2$\pm$0.2 \\
\cmidrule(l){2-11}
    & math hard & \greedy{80.0$\pm$0.5} & \greedy{79.7$\pm$0.8} & 96.7$\pm$0.1 & \greedy{80.1$\pm$0.5} & \greedy{78.9$\pm$0.7} & 95.3$\pm$0.5 & \greedy{79.5$\pm$0.8} & 78.6$\pm$0.7 & 95.5$\pm$0.3 \\
    & math fuzzy & \greedy{79.6$\pm$1.4} & 68.5$\pm$2.1 & \sample{97.6$\pm$0.3} & \greedy{78.8$\pm$0.9} & 78.1$\pm$0.8 & 95.2$\pm$0.1 & 77.9$\pm$0.9 & 75.9$\pm$1.1 & 95.7$\pm$0.3 \\
    & math soft & 76.8$\pm$1.0 & 70.8$\pm$2.5 & \sample{97.8$\pm$0.3} & 76.9$\pm$0.3 & 76.2$\pm$0.6 & 94.6$\pm$0.6 & 76.3$\pm$0.5 & 74.8$\pm$0.7 & 95.3$\pm$0.8 \\
\cmidrule(l){2-11}
    & deepscaler hard & \greedy{79.7$\pm$1.3} & \greedy{79.5$\pm$0.3} & 96.6$\pm$0.3 & \greedy{79.6$\pm$1.1} & \greedy{78.9$\pm$0.5} & 94.7$\pm$0.6 & \greedy{79.8$\pm$1.0} & 78.5$\pm$0.5 & 94.8$\pm$0.2 \\
    & deepscaler fuzzy & \greedy{78.8$\pm$0.8} & 69.6$\pm$4.7 & \sample{97.7$\pm$0.5} & 78.3$\pm$0.9 & 77.2$\pm$1.1 & 94.8$\pm$0.4 & 76.5$\pm$0.7 & 75.4$\pm$1.3 & 95.3$\pm$0.6 \\
    & deepscaler soft & \greedy{77.9$\pm$1.8} & 71.0$\pm$1.4 & \sample{98.0$\pm$0.1} & 77.5$\pm$1.3 & 76.9$\pm$1.4 & 94.4$\pm$0.6 & \greedy{77.5$\pm$1.7} & 75.5$\pm$1.6 & 95.2$\pm$0.4 \\
\midrule
\rowcolor{paleviolet}
\multirow{4}{*}{\cellcolor{white}\makecell{llama 8b\\instruct}}
    & no finetune & 82.6$\pm$0.0 & 64.9$\pm$0.0 & 98.5$\pm$0.0 & 82.6$\pm$0.0 & 69.3$\pm$0.0 & 96.1$\pm$0.0 & 81.2$\pm$0.0 & 63.8$\pm$0.0 & 96.1$\pm$0.0 \\
\cmidrule(l){2-11}
    & gsm8k hard & 81.2$\pm$0.2 & 80.6$\pm$0.4 & 95.1$\pm$0.2 & 80.0$\pm$1.3 & 69.0$\pm$0.7 & 94.5$\pm$0.3 & 78.3$\pm$2.2 & 68.3$\pm$0.7 & 94.3$\pm$0.1 \\
    & gsm8k fuzzy & \greedy{83.7$\pm$1.3} & 73.2$\pm$3.0 & \sample{98.2$\pm$0.2} & \greedy{83.4$\pm$0.7} & 73.5$\pm$1.1 & 95.5$\pm$0.3 & \greedy{82.3$\pm$0.8} & 69.8$\pm$1.7 & 95.5$\pm$0.7 \\
    & gsm8k soft & \greedy{82.6$\pm$1.6} & 73.3$\pm$3.9 & \sample{98.3$\pm$0.2} & \greedy{82.4$\pm$1.7} & 72.9$\pm$1.0 & 95.2$\pm$0.8 & \greedy{81.6$\pm$1.3} & 69.0$\pm$1.2 & 95.6$\pm$0.4 \\
\midrule
\rowcolor{paleviolet}
\multirow{4}{*}{\cellcolor{white}\makecell{qwen 3b\\instruct}}
    & no finetune & 8.9$\pm$0.0 & 17.2$\pm$0.0 & 95.1$\pm$0.0 & 10.5$\pm$0.0 & 11.5$\pm$0.0 & 58.5$\pm$0.0 & 8.6$\pm$0.0 & 13.8$\pm$0.0 & 71.9$\pm$0.0 \\
\cmidrule(l){2-11}
    & math hard & \greedy{84.0$\pm$0.8} & 82.9$\pm$0.3 & 97.2$\pm$0.3 & \greedy{84.0$\pm$0.9} & 83.2$\pm$0.3 & 94.6$\pm$0.2 & \greedy{84.0$\pm$0.7} & 83.0$\pm$0.2 & 94.9$\pm$0.2 \\
    & math fuzzy & \greedy{84.4$\pm$0.8} & 81.6$\pm$1.0 & \sample{98.1$\pm$0.2} & \greedy{84.1$\pm$0.3} & \greedy{84.2$\pm$0.3} & 94.3$\pm$0.1 & \greedy{84.0$\pm$0.6} & \greedy{84.1$\pm$0.3} & 95.0$\pm$0.3 \\
    & math soft & 82.9$\pm$0.9 & 78.7$\pm$2.6 & \sample{97.6$\pm$0.5} & 82.5$\pm$1.3 & 82.1$\pm$1.3 & 94.8$\pm$0.4 & \greedy{82.9$\pm$1.1} & 81.9$\pm$1.5 & 95.4$\pm$0.4 \\
\bottomrule\end{tabular}
}
\caption{\textbf{Results on GSM8K Test Set}.
In \greedy{blue} the best pass@1
performance and in \sample{green} the best pass@32 for each (base model,
training set) pair.} 
\label{tab:gsm8k}
\end{table}
\fi

\ifarxiv
\begin{table}[ht]
\centering\footnotesize
\resizebox{\textwidth}{!}{%
\begin{tabular}{l l|c c c|c c c|c c c}
\toprule
& & \multicolumn{9}{c}{Inference Settings}\\
Model & Training &
\makecell{Hard\\Greedy\\pass@1} &
\makecell{Hard\\Sample\\pass@1} &
\makecell{Hard\\Sample\\pass@32} &
\makecell{Fuzzy\\Greedy\\pass@1} &
\makecell{Fuzzy\\Sample\\pass@1} &
\makecell{Fuzzy\\Sample\\pass@32} &
\makecell{Soft\\Greedy\\pass@1} &
\makecell{Soft\\Sample\\pass@1} &
\makecell{Soft\\Sample\\pass@32} \\
\midrule
\rowcolor{paleviolet}
\multirow{10}{*}{\cellcolor{white}\makecell{llama 3b\\instruct}}
    & no finetune & 71.4$\pm$0.0 & 45.0$\pm$0.0 & 96.8$\pm$0.0 & 70.5$\pm$0.0 & 69.3$\pm$0.0 & 93.9$\pm$0.0 & 68.4$\pm$0.0 & 65.2$\pm$0.0 & 94.9$\pm$0.0 \\
\cmidrule(l){2-11}
    & gsm8k hard & \greedy{75.9$\pm$1.3} & 74.3$\pm$0.8 & 94.1$\pm$0.3 & 75.5$\pm$0.6 & 74.7$\pm$0.5 & 92.3$\pm$0.4 & \greedy{75.7$\pm$0.5} & 74.2$\pm$0.4 & 92.6$\pm$0.1 \\
    & gsm8k fuzzy & \greedy{76.7$\pm$1.8} & 66.4$\pm$2.4 & \sample{97.4$\pm$0.3} & \greedy{76.4$\pm$2.1} & \greedy{75.2$\pm$1.8} & 92.0$\pm$1.1 & \greedy{75.1$\pm$1.8} & 73.5$\pm$1.7 & 93.5$\pm$0.8 \\
    & gsm8k soft & \greedy{77.2$\pm$0.9} & 70.5$\pm$3.3 & \sample{97.9$\pm$0.3} & \greedy{76.8$\pm$0.8} & \greedy{76.0$\pm$1.4} & 93.4$\pm$0.3 & 74.5$\pm$1.5 & 74.1$\pm$2.0 & 94.2$\pm$0.2 \\
\cmidrule(l){2-11}
    & math hard & \greedy{80.0$\pm$0.5} & \greedy{79.7$\pm$0.8} & 96.7$\pm$0.1 & \greedy{80.1$\pm$0.5} & \greedy{78.9$\pm$0.7} & 95.3$\pm$0.5 & \greedy{79.5$\pm$0.8} & 78.6$\pm$0.7 & 95.5$\pm$0.3 \\
    & math fuzzy & \greedy{79.6$\pm$1.4} & 68.5$\pm$2.1 & \sample{97.6$\pm$0.3} & \greedy{78.8$\pm$0.9} & 78.1$\pm$0.8 & 95.2$\pm$0.1 & 77.9$\pm$0.9 & 75.9$\pm$1.1 & 95.7$\pm$0.3 \\
    & math soft & 76.8$\pm$1.0 & 70.8$\pm$2.5 & \sample{97.8$\pm$0.3} & 76.9$\pm$0.3 & 76.2$\pm$0.6 & 94.6$\pm$0.6 & 76.3$\pm$0.5 & 74.8$\pm$0.7 & 95.3$\pm$0.8 \\
\cmidrule(l){2-11}
    & deepscaler hard & \greedy{79.7$\pm$1.3} & \greedy{79.5$\pm$0.3} & 96.6$\pm$0.3 & \greedy{79.6$\pm$1.1} & \greedy{78.9$\pm$0.5} & 94.7$\pm$0.6 & \greedy{79.8$\pm$1.0} & 78.5$\pm$0.5 & 94.8$\pm$0.2 \\
    & deepscaler fuzzy & \greedy{78.8$\pm$0.8} & 69.6$\pm$4.7 & \sample{97.7$\pm$0.5} & 78.3$\pm$0.9 & 77.2$\pm$1.1 & 94.8$\pm$0.4 & 76.5$\pm$0.7 & 75.4$\pm$1.3 & 95.3$\pm$0.6 \\
    & deepscaler soft & \greedy{77.9$\pm$1.8} & 71.0$\pm$1.4 & \sample{98.0$\pm$0.1} & 77.5$\pm$1.3 & 76.9$\pm$1.4 & 94.4$\pm$0.6 & \greedy{77.5$\pm$1.7} & 75.5$\pm$1.6 & 95.2$\pm$0.4 \\
\midrule
\rowcolor{paleviolet}
\multirow{4}{*}{\cellcolor{white}\makecell{llama 8b\\instruct}}
    & no finetune & 82.6$\pm$0.0 & 64.9$\pm$0.0 & 98.5$\pm$0.0 & 82.6$\pm$0.0 & 69.3$\pm$0.0 & 96.1$\pm$0.0 & 81.2$\pm$0.0 & 63.8$\pm$0.0 & 96.1$\pm$0.0 \\
\cmidrule(l){2-11}
    & gsm8k hard & 81.2$\pm$0.2 & 80.6$\pm$0.4 & 95.1$\pm$0.2 & 80.0$\pm$1.3 & 69.0$\pm$0.7 & 94.5$\pm$0.3 & 78.3$\pm$2.2 & 68.3$\pm$0.7 & 94.3$\pm$0.1 \\
    & gsm8k fuzzy & \greedy{83.7$\pm$1.3} & 73.2$\pm$3.0 & \sample{98.2$\pm$0.2} & \greedy{83.4$\pm$0.7} & 73.5$\pm$1.1 & 95.5$\pm$0.3 & \greedy{82.3$\pm$0.8} & 69.8$\pm$1.7 & 95.5$\pm$0.7 \\
    & gsm8k soft & \greedy{82.6$\pm$1.6} & 73.3$\pm$3.9 & \sample{98.3$\pm$0.2} & \greedy{82.4$\pm$1.7} & 72.9$\pm$1.0 & 95.2$\pm$0.8 & \greedy{81.6$\pm$1.3} & 69.0$\pm$1.2 & 95.6$\pm$0.4 \\
\midrule
\rowcolor{paleviolet}
\multirow{4}{*}{\cellcolor{white}\makecell{qwen 3b\\instruct}}
    & no finetune & 8.9$\pm$0.0 & 17.2$\pm$0.0 & 95.1$\pm$0.0 & 10.5$\pm$0.0 & 11.5$\pm$0.0 & 58.5$\pm$0.0 & 8.6$\pm$0.0 & 13.8$\pm$0.0 & 71.9$\pm$0.0 \\
\cmidrule(l){2-11}
    & math hard & \greedy{84.0$\pm$0.8} & 82.9$\pm$0.3 & 97.2$\pm$0.3 & \greedy{84.0$\pm$0.9} & 83.2$\pm$0.3 & 94.6$\pm$0.2 & \greedy{84.0$\pm$0.7} & 83.0$\pm$0.2 & 94.9$\pm$0.2 \\
    & math fuzzy & \greedy{84.4$\pm$0.8} & 81.6$\pm$1.0 & \sample{98.1$\pm$0.2} & \greedy{84.1$\pm$0.3} & \greedy{84.2$\pm$0.3} & 94.3$\pm$0.1 & \greedy{84.0$\pm$0.6} & \greedy{84.1$\pm$0.3} & 95.0$\pm$0.3 \\
    & math soft & 82.9$\pm$0.9 & 78.7$\pm$2.6 & \sample{97.6$\pm$0.5} & 82.5$\pm$1.3 & 82.1$\pm$1.3 & 94.8$\pm$0.4 & \greedy{82.9$\pm$1.1} & 81.9$\pm$1.5 & 95.4$\pm$0.4 \\
\bottomrule\end{tabular}
}
\caption{\textbf{Results on GSM8K test set}.
In \greedy{blue} the best pass@1
performance and in \sample{green} the best pass@32 for each (base model,
training set) pair.} 
\label{tab:gsm8k}
\end{table}
\fi
\begin{table}[ht]\centering\footnotesize
\resizebox{\textwidth}{!}{%
\begin{tabular}{l l|c c c|c c c|c c c}
\toprule
& & \multicolumn{9}{c}{Inference Settings}\\
Model & Training &
\makecell{Hard\\Greedy\\pass@1} &
\makecell{Hard\\Sample\\pass@1} &
\makecell{Hard\\Sample\\pass@32} &
\makecell{Fuzzy\\Greedy\\pass@1} &
\makecell{Fuzzy\\Sample\\pass@1} &
\makecell{Fuzzy\\Sample\\pass@32} &
\makecell{Soft\\Greedy\\pass@1} &
\makecell{Soft\\Sample\\pass@1} &
\makecell{Soft\\Sample\\pass@32} \\

\midrule
\rowcolor{paleviolet}
\multirow{10}{*}{\cellcolor{white}\makecell{llama 3b\\instruct}}
    & no finetune & 38.0$\pm$0.0 & 25.2$\pm$0.0 & 82.0$\pm$0.0 & 37.8$\pm$0.0 & 34.9$\pm$0.0 & 78.4$\pm$0.0 & 37.2$\pm$0.0 & 33.1$\pm$0.0 & 79.2$\pm$0.0 \\
\cmidrule(l){2-11}
    & gsm8k hard & 34.6$\pm$0.2 & 31.2$\pm$0.4 & 72.4$\pm$1.1 & 34.6$\pm$0.0 & 32.7$\pm$0.5 & 71.6$\pm$0.9 & 33.2$\pm$1.6 & 32.1$\pm$0.8 & 70.1$\pm$0.2 \\
    & gsm8k fuzzy & \greedy{42.6$\pm$3.2} & 32.6$\pm$1.6 & 83.2$\pm$1.1 & \greedy{41.6$\pm$2.1} & 41.2$\pm$2.0 & 79.5$\pm$1.7 & 40.3$\pm$1.5 & 38.7$\pm$1.7 & 79.5$\pm$2.0 \\
    & gsm8k soft & \greedy{43.5$\pm$1.4} & 37.0$\pm$1.0 & \sample{84.8$\pm$0.4} & \greedy{44.2$\pm$0.5} & 42.5$\pm$0.7 & 78.1$\pm$0.4 & \greedy{42.0$\pm$2.3} & 40.7$\pm$0.8 & 79.6$\pm$1.0 \\
\cmidrule(l){2-11}
    & math hard & \greedy{49.1$\pm$1.7} & \greedy{47.9$\pm$0.9} & 78.7$\pm$0.9 & \greedy{49.7$\pm$1.0} & 47.0$\pm$1.0 & 76.5$\pm$0.2 & \greedy{48.1$\pm$1.7} & 46.8$\pm$1.0 & 76.3$\pm$1.0 \\
    & math fuzzy & \greedy{48.5$\pm$0.3} & 37.5$\pm$1.6 & \sample{81.5$\pm$0.7} & 47.4$\pm$0.7 & 46.9$\pm$0.9 & 79.7$\pm$0.6 & 45.6$\pm$1.6 & 44.8$\pm$0.6 & 79.9$\pm$0.1 \\
    & math soft & 44.5$\pm$0.6 & 39.1$\pm$1.8 & \sample{82.2$\pm$1.6} & 45.1$\pm$1.4 & 43.4$\pm$0.6 & \sample{79.3$\pm$1.7} & 43.2$\pm$0.8 & 42.0$\pm$0.3 & \sample{80.2$\pm$0.9} \\
\cmidrule(l){2-11}
    & deepscaler hard & \greedy{49.6$\pm$0.9} & 48.3$\pm$0.6 & 78.1$\pm$0.7 & \greedy{49.2$\pm$0.7} & 47.9$\pm$0.4 & 77.7$\pm$0.5 & \greedy{50.0$\pm$0.6} & 47.4$\pm$0.2 & 77.7$\pm$1.1 \\
    & deepscaler fuzzy & 46.5$\pm$2.2 & 38.5$\pm$2.6 & \sample{83.3$\pm$0.3} & 46.0$\pm$1.7 & 45.3$\pm$1.2 & 79.8$\pm$0.3 & 45.8$\pm$0.9 & 43.5$\pm$1.4 & 78.9$\pm$0.5 \\
    & deepscaler soft & 44.7$\pm$1.2 & 37.4$\pm$1.3 & 80.6$\pm$2.2 & 44.6$\pm$1.1 & 42.8$\pm$0.7 & 78.5$\pm$1.2 & 40.9$\pm$1.4 & 41.3$\pm$1.6 & 77.9$\pm$1.2 \\
\midrule
\rowcolor{paleviolet}
\multirow{4}{*}{\cellcolor{white}\makecell{llama 8b\\instruct}}
    & no finetune & 44.4$\pm$0.0 & 31.1$\pm$0.0 & 79.8$\pm$0.0 & 45.2$\pm$0.0 & 32.2$\pm$0.0 & 78.6$\pm$0.0 & 42.4$\pm$0.0 & 29.3$\pm$0.0 & 73.6$\pm$0.0 \\
\cmidrule(l){2-11}
    & gsm8k hard & 20.2$\pm$0.8 & 19.8$\pm$1.2 & 45.4$\pm$3.2 & 20.8$\pm$1.0 & 9.7$\pm$0.6 & 32.3$\pm$0.2 & 20.0$\pm$1.8 & 9.7$\pm$0.6 & 32.2$\pm$0.4 \\
    & gsm8k fuzzy & \greedy{44.6$\pm$2.1} & 33.7$\pm$2.5 & \sample{83.1$\pm$0.9} & \greedy{45.1$\pm$1.4} & 33.9$\pm$0.9 & 75.9$\pm$2.1 & \greedy{43.7$\pm$0.8} & 31.1$\pm$1.2 & 75.7$\pm$2.1 \\
    & gsm8k soft & \greedy{44.7$\pm$2.3} & 34.8$\pm$2.2 & \sample{83.9$\pm$1.1} & \greedy{44.3$\pm$1.7} & 33.9$\pm$0.7 & 76.9$\pm$1.4 & \greedy{44.3$\pm$0.9} & 31.0$\pm$0.5 & 75.4$\pm$1.1 \\
\midrule
\rowcolor{paleviolet}
\multirow{4}{*}{\cellcolor{white}\makecell{qwen 3b\\instruct}}
    & no finetune & 29.0$\pm$0.0 & 25.5$\pm$0.0 & 81.0$\pm$0.0 & 27.6$\pm$0.0 & 28.1$\pm$0.0 & 69.0$\pm$0.0 & 27.0$\pm$0.0 & 29.6$\pm$0.0 & 73.6$\pm$0.0 \\
\cmidrule(l){2-11}
    & math hard & \greedy{59.0$\pm$1.7} & 57.1$\pm$0.2 & \sample{83.6$\pm$1.0} & \greedy{58.9$\pm$1.9} & 57.5$\pm$0.4 & 79.9$\pm$1.0 & \greedy{58.3$\pm$0.2} & 57.1$\pm$0.5 & 80.2$\pm$0.7 \\
    & math fuzzy & \greedy{58.1$\pm$0.9} & 55.5$\pm$1.2 & \sample{84.4$\pm$0.2} & \greedy{58.4$\pm$0.8} & \greedy{57.9$\pm$0.8} & 79.1$\pm$0.7 & \greedy{58.0$\pm$1.4} & \greedy{57.7$\pm$1.0} & 80.3$\pm$0.8 \\
    & math soft & 54.7$\pm$0.3 & 52.2$\pm$1.1 & \sample{84.4$\pm$0.7} & 54.9$\pm$0.5 & 54.3$\pm$0.1 & 79.1$\pm$0.5 & 54.7$\pm$1.2 & 54.6$\pm$0.6 & 80.3$\pm$1.5 \\
\bottomrule\end{tabular}
}
\caption{\textbf{Results on MATH-500}.
In \greedy{blue} the best pass@1
performance and in \sample{green} the best pass@32 for each (base model,
training set) pair. } 
\label{tab:math500}
\end{table}

\begin{table}[ht]\centering\footnotesize
\resizebox{\textwidth}{!}{%
\begin{tabular}{l l|c c c|c c c|c c c}
\toprule
& & \multicolumn{9}{c}{Inference Settings}\\
Model & Training &
\makecell{Hard\\Greedy\\pass@1} &
\makecell{Hard\\Sample\\pass@1} &
\makecell{Hard\\Sample\\pass@32} &
\makecell{Fuzzy\\Greedy\\pass@1} &
\makecell{Fuzzy\\Sample\\pass@1} &
\makecell{Fuzzy\\Sample\\pass@32} &
\makecell{Soft\\Greedy\\pass@1} &
\makecell{Soft\\Sample\\pass@1} &
\makecell{Soft\\Sample\\pass@32} \\
\midrule
\rowcolor{paleviolet}
\multirow{10}{*}{\cellcolor{white}\makecell{llama 3b\\instruct}}
    & no finetune & 17.9$\pm$0.0 & 12.0$\pm$0.0 & 52.3$\pm$0.0 & 18.4$\pm$0.0 & 15.4$\pm$0.0 & 53.2$\pm$0.0 & 14.4$\pm$0.0 & 14.7$\pm$0.0 & 52.4$\pm$0.0 \\
\cmidrule(l){2-11}
    & gsm8k hard & 10.6$\pm$0.9 & 9.4$\pm$0.5 & 39.3$\pm$1.3 & 11.0$\pm$1.2 & 10.1$\pm$0.4 & 39.1$\pm$0.8 & 11.1$\pm$0.5 & 9.8$\pm$0.3 & 38.0$\pm$1.9 \\
    & gsm8k fuzzy & 17.4$\pm$0.9 & 13.0$\pm$1.1 & 55.2$\pm$1.2 & \greedy{17.6$\pm$1.3} & 16.4$\pm$1.1 & \sample{55.3$\pm$2.1} & 15.5$\pm$0.3 & 15.5$\pm$0.6 & 55.2$\pm$1.5 \\
    & gsm8k soft & \greedy{18.9$\pm$0.2} & 15.5$\pm$0.7 & \sample{58.6$\pm$1.6} & \greedy{19.3$\pm$1.2} & 18.2$\pm$0.4 & \sample{57.5$\pm$0.8} & 17.4$\pm$0.4 & 16.9$\pm$0.2 & \sample{57.5$\pm$0.7} \\
\cmidrule(l){2-11}
    & math hard & \greedy{22.7$\pm$1.1} & \greedy{22.2$\pm$0.4} & 49.3$\pm$1.3 & \greedy{22.2$\pm$1.2} & \greedy{22.1$\pm$0.3} & 48.5$\pm$1.5 & \greedy{22.7$\pm$0.7} & \greedy{21.9$\pm$0.4} & 48.0$\pm$1.4 \\
    & math fuzzy & \greedy{21.6$\pm$0.6} & 15.9$\pm$0.8 & 52.8$\pm$1.6 & \greedy{21.2$\pm$1.1} & 21.0$\pm$0.7 & \sample{55.4$\pm$1.7} & 19.9$\pm$0.4 & 19.6$\pm$0.5 & \sample{56.2$\pm$0.6} \\
    & math soft & 19.2$\pm$1.1 & 16.6$\pm$0.4 & \sample{55.8$\pm$1.2} & 20.2$\pm$0.5 & 19.5$\pm$0.4 & 53.5$\pm$2.1 & 19.9$\pm$0.7 & 18.2$\pm$0.3 & 53.9$\pm$1.1 \\
\cmidrule(l){2-11}
    & deepscaler hard & \greedy{23.2$\pm$1.4} & \greedy{23.8$\pm$0.6} & 50.8$\pm$1.5 & \greedy{23.0$\pm$0.6} & \greedy{23.7$\pm$0.3} & 49.4$\pm$1.1 & \greedy{24.0$\pm$1.2} & \greedy{23.3$\pm$0.5} & 49.0$\pm$1.9 \\
    & deepscaler fuzzy & 19.5$\pm$1.0 & 16.6$\pm$1.4 & \sample{54.4$\pm$2.1} & 20.7$\pm$1.9 & 19.7$\pm$0.9 & \sample{53.1$\pm$1.0} & 20.5$\pm$1.1 & 18.5$\pm$1.1 & \sample{53.8$\pm$1.3} \\
    & deepscaler soft & 21.0$\pm$0.6 & 16.4$\pm$0.7 & \sample{53.3$\pm$1.7} & 18.9$\pm$1.3 & 19.5$\pm$0.1 & \sample{53.2$\pm$1.5} & 18.4$\pm$0.3 & 18.0$\pm$0.4 & \sample{51.5$\pm$1.6} \\
\midrule
\rowcolor{paleviolet}
\multirow{4}{*}{\cellcolor{white}\makecell{llama 8b\\instruct}}
    & no finetune & 19.6$\pm$0.0 & 11.7$\pm$0.0 & 52.6$\pm$0.0 & 18.7$\pm$0.0 & 12.8$\pm$0.0 & 46.4$\pm$0.0 & 15.7$\pm$0.0 & 11.5$\pm$0.0 & 48.3$\pm$0.0 \\
\cmidrule(l){2-11}
    & gsm8k hard & 3.8$\pm$0.5 & 3.4$\pm$0.5 & 16.4$\pm$2.4 & 3.8$\pm$0.8 & 1.2$\pm$0.2 & 7.6$\pm$0.4 & 3.9$\pm$0.1 & 1.0$\pm$0.2 & 6.7$\pm$0.8 \\
    & gsm8k fuzzy & \greedy{18.0$\pm$1.4} & 12.5$\pm$1.3 & \sample{56.0$\pm$2.6} & \greedy{17.9$\pm$1.3} & 14.1$\pm$1.3 & 52.5$\pm$2.9 & \greedy{18.1$\pm$1.4} & 13.0$\pm$1.4 & 51.5$\pm$2.7 \\
    & gsm8k soft & \greedy{17.9$\pm$1.0} & 13.1$\pm$0.7 & \sample{56.8$\pm$1.2} & \greedy{18.7$\pm$1.3} & 13.8$\pm$0.4 & 53.0$\pm$1.5 & \greedy{17.1$\pm$0.6} & 13.0$\pm$0.5 & 51.1$\pm$0.9 \\
\midrule
\rowcolor{paleviolet}
\multirow{4}{*}{\cellcolor{white}\makecell{qwen 3b\\instruct}}
    & no finetune & 16.9$\pm$0.0 & 14.3$\pm$0.0 & 50.2$\pm$0.0 & 17.3$\pm$0.0 & 16.4$\pm$0.0 & 40.6$\pm$0.0 & 17.3$\pm$0.0 & 16.7$\pm$0.0 & 45.3$\pm$0.0 \\
\cmidrule(l){2-11}
    & math hard & \greedy{29.4$\pm$0.3} & 27.8$\pm$0.5 & \sample{61.1$\pm$0.6} & \greedy{29.6$\pm$0.8} & 28.6$\pm$0.4 & 57.2$\pm$0.4 & \greedy{29.1$\pm$0.7} & 28.3$\pm$0.6 & 55.7$\pm$0.5 \\
    & math fuzzy & 27.2$\pm$0.5 & 24.4$\pm$0.7 & \sample{60.7$\pm$0.5} & 27.4$\pm$1.2 & 27.1$\pm$0.5 & 54.0$\pm$0.1 & 27.2$\pm$0.7 & 26.6$\pm$0.7 & 55.4$\pm$0.1 \\
    & math soft & 24.3$\pm$1.8 & 22.0$\pm$0.5 & 58.5$\pm$1.0 & 22.8$\pm$1.5 & 23.9$\pm$0.6 & 53.2$\pm$2.9 & 23.1$\pm$2.3 & 23.5$\pm$0.4 & 54.9$\pm$1.7 \\
\bottomrule\end{tabular}
}
\caption{\textbf{Results on OlympiadBench}.
In \greedy{blue} the best pass@1
performance and in \sample{green} the best pass@32 for each (base model,
training set) pair.} 
\label{tab:olympiad}
\end{table}

\section{Ablations}\label{app:ablations}
All ablations were run with two independent random seeds per experiment.

\begin{figure}
  \centering
  \includegraphics[width=.45\textwidth]{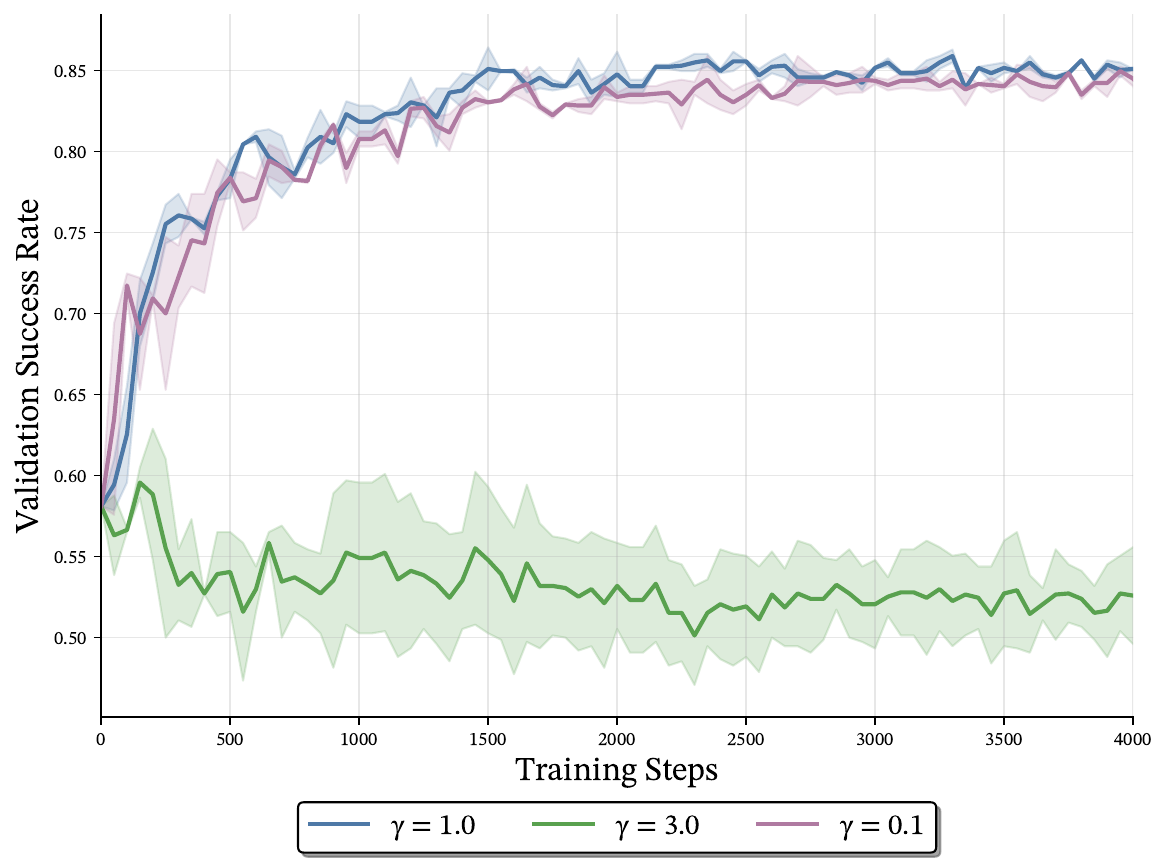}%
  \includegraphics[width=.45\textwidth]{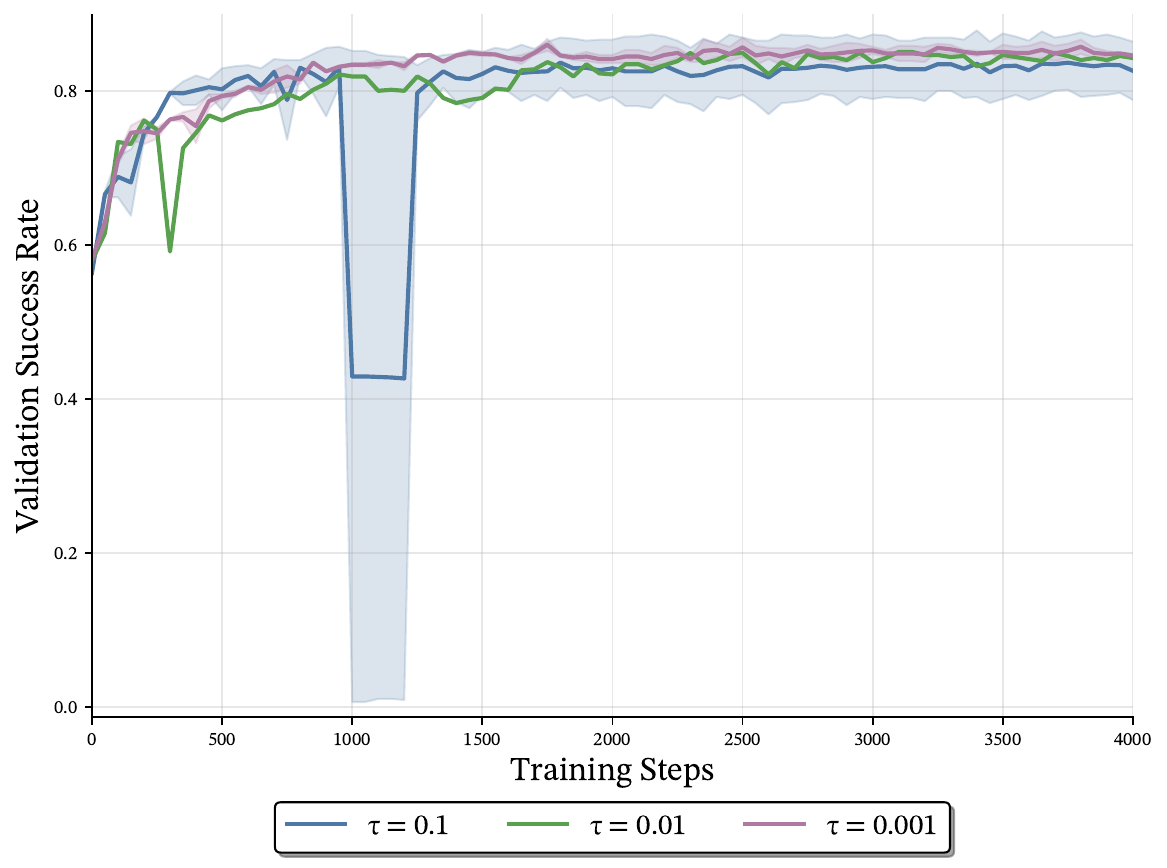}
  \caption{\textbf{Validation performance for: (a) noise scale ablation, (b) temperature ablation}, on Llama 3B Instruct trained with fuzzy models on GSM8K. \emph{Fuzzy training appears robust to noise scale factors 0.1-1.0 and temperature values 0.1-0.0001}.}
  \label{fig:ablations}
\end{figure}

\subsection{Noise Scale}\label{app:ablations:noise-scale}
\begin{table}[ht]\centering\footnotesize
\resizebox{\textwidth}{!}{%
\begin{tabular}{l|c c c |c c c |c c c}
\toprule
 & \multicolumn{9}{c}{Inference setting} \\
Noise Factor & \makecell{Hard\\Greedy\\pass@1} &
\makecell{Hard\\Sample\\pass@1} &
\makecell{Hard\\Sample\\pass@32} &
\makecell{Fuzzy\\Greedy\\pass@1} &
\makecell{Fuzzy\\Sample\\pass@1} &
\makecell{Fuzzy\\Sample\\pass@32} &
\makecell{Soft\\Greedy\\pass@1} &
\makecell{Soft\\Sample\\pass@1} &
\makecell{Soft\\Sample\\pass@32} \\
\midrule
$\gamma$=0.1 & 75.5$\pm$0.8 & 68.9$\pm$3.2 & 97.5$\pm$0.1 & 74.5$\pm$0.9 & 74.5$\pm$1.1 & 92.9$\pm$0.3 & 73.5$\pm$1.2 & 72.9$\pm$1.5 & 93.9$\pm$0.2 \\
$\gamma$=0.33 & 76.7$\pm$1.8 & 66.4$\pm$2.4 & 97.4$\pm$0.3 & 76.4$\pm$2.1 & 75.2$\pm$1.8 & 92.0$\pm$1.1 & 75.1$\pm$1.8 & 73.5$\pm$1.7 & 93.5$\pm$0.8 \\
$\gamma$=1.0 & 78.1$\pm$0.2 & 71.7$\pm$0.1 & 97.7$\pm$0.2 & 77.7$\pm$0.0 & 77.0$\pm$0.1 & 93.3$\pm$0.5 & 77.5$\pm$0.5 & 75.8$\pm$0.3 & 94.4$\pm$0.6 \\
$\gamma$=3.0 & 65.4$\pm$1.9 & 37.8$\pm$4.4 & 95.9$\pm$0.6 & 65.1$\pm$1.7 & 63.1$\pm$2.1 & 91.1$\pm$1.3 & 60.5$\pm$2.9 & 57.3$\pm$3.0 & 93.1$\pm$0.6 \\
\bottomrule\end{tabular}
}
\caption{\textbf{Results of noise factor ablation study} on GSM8K test set of  for Llama 3B Instruct trained with fuzzy model on GSM8K train. \emph{Fuzzy training performance is relatively robust to scale factors less than or equal to 1.0.}}
\label{tab:scale}
\end{table}

In Table \ref{tab:scale}, we report test performance on GSM8K test set
for Llama 3b Instruct trained with the fuzzy RLOO, varying the scale
factor, $\gamma$,  applied to the root mean square embedding norm to
compute the noise scale $\sigma$. Further, in Figure \ref{fig:ablations},
we report the greedy validation performance on GSM8K train set. We
observe that our algorithm appears robust to scale factors 1 and below;
whereas for $\gamma=3$, there is a collapse in learning as noise gets too
large.

\subsection{Noise Placement}\label{app:ablations:noise-placement}
\begin{table}[ht]\centering\footnotesize
\resizebox{\textwidth}{!}{%
\begin{tabular}{l l l l| c c c| c c c|c c c}
\toprule
 &  & & & \multicolumn{9}{c}{Inference setting} \\
Training & Noise Loc & Temp & Top-k & \makecell{Hard\\Greedy\\pass@1} &
\makecell{Hard\\Sample\\pass@1} &
\makecell{Hard\\Sample\\pass@32} &
\makecell{Fuzzy\\Greedy\\pass@1} &
\makecell{Fuzzy\\Sample\\pass@1} &
\makecell{Fuzzy\\Sample\\pass@32} &
\makecell{Soft\\Greedy\\pass@1} &
\makecell{Soft\\Sample\\pass@1} &
\makecell{Soft\\Sample\\pass@32} \\
\midrule
Training None & Tokens & 1.0 & 1 & 71.0$\pm$0.0 & 45.0$\pm$0.0 & 96.8$\pm$0.0 & 70.8$\pm$0.0 & 69.3$\pm$0.0 & 93.9$\pm$0.0 & 67.7$\pm$0.0 & 65.2$\pm$0.0 & 94.9$\pm$0.0 \\
Fuzzy & Embedding & 0.0001 & - & 76.7$\pm$1.8 & 66.4$\pm$2.4 & 97.4$\pm$0.3 & 76.4$\pm$2.1 & 75.2$\pm$1.8 & 92.0$\pm$1.1 & 75.1$\pm$1.8 & 73.5$\pm$1.7 & 93.5$\pm$0.8 \\
Soft & Embedding & 0.5 & - & 77.2$\pm$0.9 & 70.6$\pm$3.4 & 97.9$\pm$0.3 & 76.8$\pm$0.8 & 76.0$\pm$1.3 & 93.4$\pm$0.3 & 74.5$\pm$1.5 & 74.1$\pm$2.1 & 94.2$\pm$0.2 \\
\cmidrule(l){1-13}
Fuzzy & Final Hidden & 0.0001 & - & 66.2$\pm$4.1 & 46.6$\pm$12.4 & 96.1$\pm$1.0 & 66.0$\pm$3.5 & 64.6$\pm$4.1 & 89.2$\pm$0.9 & 62.7$\pm$4.6 & 60.1$\pm$5.8 & 90.6$\pm$0.6 \\
Soft & Final Hidden & 0.5 & - & 68.0$\pm$3.3 & 40.0$\pm$10.0 & 94.5$\pm$1.7 & 66.8$\pm$3.5 & 65.1$\pm$3.6 & 89.3$\pm$1.0 & 64.3$\pm$3.3 & 61.0$\pm$3.7 & 91.7$\pm$0.6 \\
Fuzzy & Logits & 0.0001 & - & 66.8$\pm$2.4 & 32.3$\pm$2.7 & 94.3$\pm$0.8 & 65.2$\pm$3.5 & 63.9$\pm$2.9 & 91.3$\pm$2.2 & 61.1$\pm$4.2 & 56.8$\pm$4.3 & 94.0$\pm$1.4 \\
Soft & Logits & 0.5 & - & 60.0$\pm$14.3 & 35.8$\pm$12.6 & 94.3$\pm$2.1 & 59.0$\pm$13.6 & 58.6$\pm$12.9 & 90.0$\pm$3.3 & 56.8$\pm$13.0 & 53.8$\pm$13.4 & 92.4$\pm$2.6 \\
Soft & Logits & 1.0 & 5 & 72.8$\pm$0.1 & 68.5$\pm$0.7 & 94.0$\pm$0.7 & 72.1$\pm$0.2 & 71.5$\pm$0.6 & 90.9$\pm$0.5 & 71.6$\pm$0.6 & 70.7$\pm$0.6 & 91.5$\pm$0.4 \\
Soft & Logits & 1.0 & 50 & 74.2$\pm$0.2 & 58.4$\pm$5.8 & 97.5$\pm$0.6 & 73.4$\pm$0.5 & 71.9$\pm$0.1 & 94.2$\pm$0.0 & 72.0$\pm$0.3 & 68.7$\pm$0.3 & 94.7$\pm$0.3 \\
\bottomrule\end{tabular}
}
\caption{\textbf{Results of noise placement ablation study } on GSM8K test set for Llama 3B Instruct trained with fuzzy/soft models on GSM8K train. \emph{We only see evidence of increasing performance on some metrics for top-k=5 and top-k=50. Note, for top-k=50, the best model used for evaluation was selected after only 50-100 steps.}}
\label{tab:placement}
\end{table}
\begin{figure}
  \centering
  \includegraphics[width=0.45\textwidth]{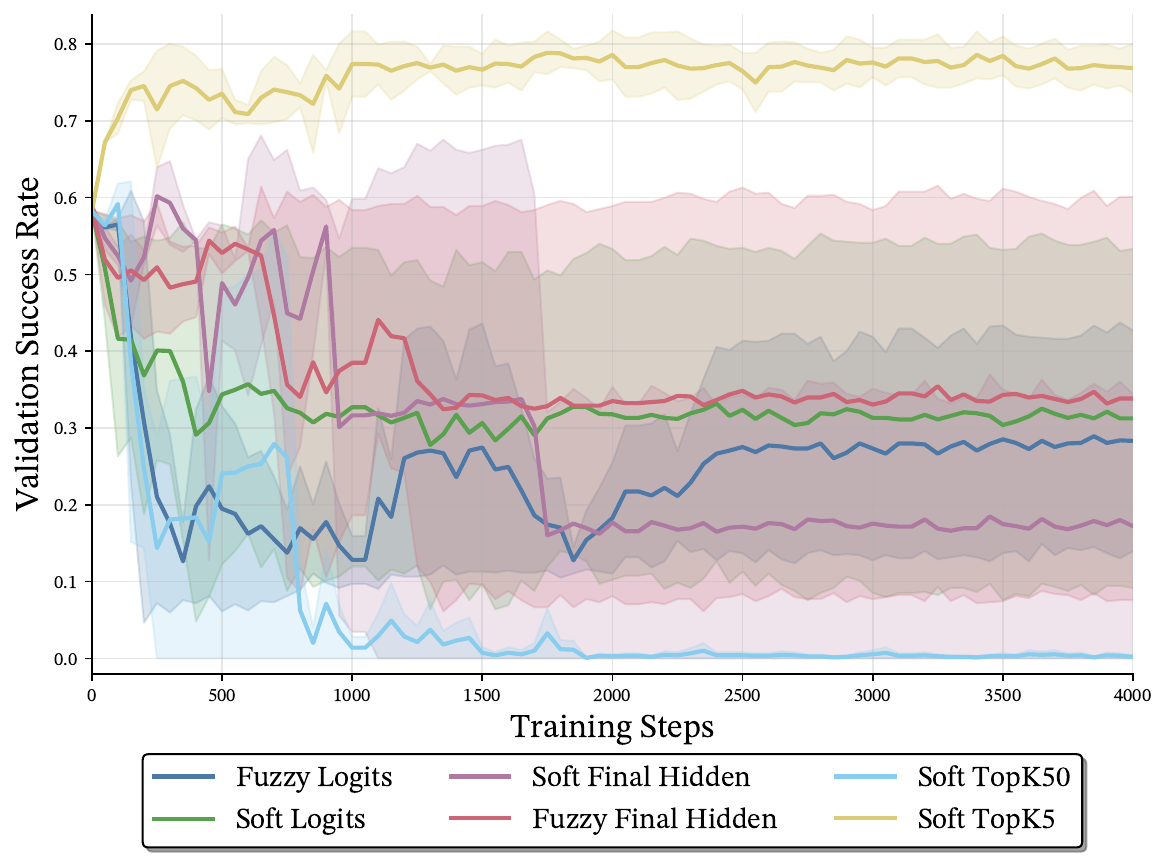}%
  \caption{\textbf{Validation performance for noise placement ablation} on Llama 3B Instruct trained with fuzzy/soft models on GSM8K. \emph{We tried placing the noise on the (top-k) logits and final hidden layer outputs; only placing noise on the top-k=5 logits shows signs of learning.}}
  \label{fig:noise_placement_ablations}
\end{figure}
In Table \ref{tab:placement}, we report test performance on GSM8K test set for Llama 3b Instruct trained with the fuzzy and soft RLOO, varying the placement of noise in our model. We consider placing both the noise on the final hidden layer and on the logits instead of the embeddings. We pose that the dimensionality of the noise on the logits layer may be too high for learning given the signal to noise ratio and so consider a top-k variant where we only add noise to the top-k logits and set the remaining logits to negative infinity. In Figure \ref{fig:noise_placement_ablations}, we report the greedy validation performance on GSM8K train set. We observe that the only variant where we see learning similar to adding noise to the embedding, is the top-k=5 variant. Interestingly, this is less reflected in the test performance in Table \ref{tab:placement}, where only some metrics show improved performance compared to the base model; however, note that the inference settings are as described in Section \ref{sec:experiments} and potential gains may be seen through soft inference that also adds noise to the top-k logits instead of the embeddings. Despite a collapse in validation performance, the top-k 50 test performance is on average higher than the base model suggesting some learning, however, on inspection the model checkpoints with the best validation performance which are used for evaluation were those after 50-100 steps.

\subsection{Temperature}\label{app:ablations:temperature}
\begin{table}[ht]\centering\footnotesize
\resizebox{\textwidth}{!}{%
\begin{tabular}{l |c c c| c c c| c c c}
\toprule
Temperature & \multicolumn{9}{c}{Inference setting} \\
& \makecell{Hard\\Greedy\\pass@1} &
\makecell{Hard\\Sample\\pass@1} &
\makecell{Hard\\Sample\\pass@32} &
\makecell{Fuzzy\\Greedy\\pass@1} &
\makecell{Fuzzy\\Sample\\pass@1} &
\makecell{Fuzzy\\Sample\\pass@32} &
\makecell{Soft\\Greedy\\pass@1} &
\makecell{Soft\\Sample\\pass@1} &
\makecell{Soft\\Sample\\pass@32} \\
\specialrule{1.0pt}{0pt}{0pt}

$\tau$=0.0001 & 76.7$\pm$1.8 & 66.4$\pm$2.4 & 97.4$\pm$0.3 & 76.4$\pm$2.1 & 75.2$\pm$1.8 & 92.0$\pm$1.1 & 75.1$\pm$1.8 & 73.5$\pm$1.7 & 93.5$\pm$0.8 \\
$\tau$=0.001 & 77.4$\pm$0.3 & 67.8$\pm$1.5 & 97.8$\pm$0.2 & 77.3$\pm$0.1 & 76.2$\pm$0.4 & 92.2$\pm$0.2 & 76.5$\pm$1.2 & 74.6$\pm$0.1 & 93.9$\pm$0.6 \\
$\tau$=0.01 & 76.9$\pm$0.2 & 70.7$\pm$1.1 & 97.3$\pm$0.0 & 77.2$\pm$0.1 & 76.2$\pm$0.4 & 93.7$\pm$0.1 & 75.5$\pm$0.2 & 74.6$\pm$0.5 & 94.2$\pm$0.3 \\
$\tau$=0.1 & 76.3$\pm$2.0 & 63.1$\pm$7.5 & 97.4$\pm$0.2 & 75.1$\pm$2.3 & 74.5$\pm$2.5 & 94.1$\pm$2.5 & 74.3$\pm$3.3 & 72.4$\pm$2.7 & 94.4$\pm$2.4 \\
\bottomrule\end{tabular}
}
\caption{\textbf{Results of temperature ablation study} on GSM8K test set for Llama 3B Instruct trained with fuzzy model on GSM8K train. \emph{Fuzzy training performance is comparable across temperature values 0.1-0.0001}.}
\label{tab:temp}
\end{table}
In Table \ref{tab:temp}, we report test performance on GSM8K test set for Llama 3b Instruct trained with the fuzzy RLOO, varying the temperature, $\tau$. Further, in Figure \ref{fig:ablations}, we report the greedy validation performance on GSM8K train set. We observe that our algorithm appears robust to temperatures between 0.1 and 0.0001.

\section{Supplementary Results}\label{app:supplementary-results}
\subsection{Training and Validation Performances}\label{app:supplementary-results:training}

\begin{figure}
    \centering
    \includegraphics[width=0.95\linewidth]{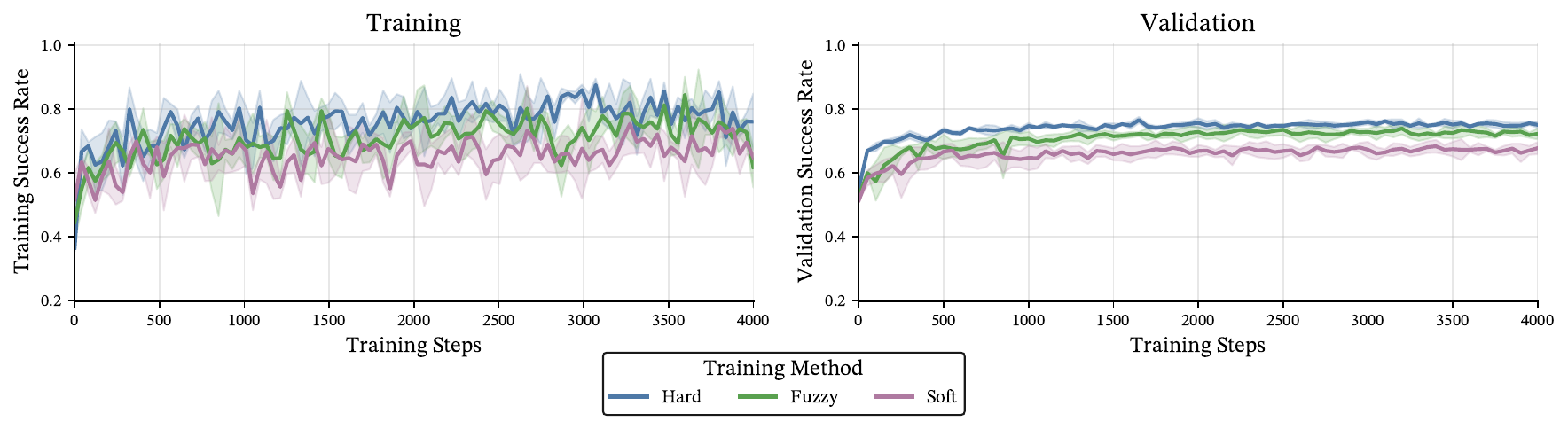}
    \caption{\textbf{Llama 3b Instruct trained on MATH train} (a) Training performance across steps. (b) Greedy validation performance used for model selection.}
    \label{fig:training_llama_3b_math}
\end{figure}

\begin{figure}
    \centering
    \includegraphics[width=0.95\linewidth]{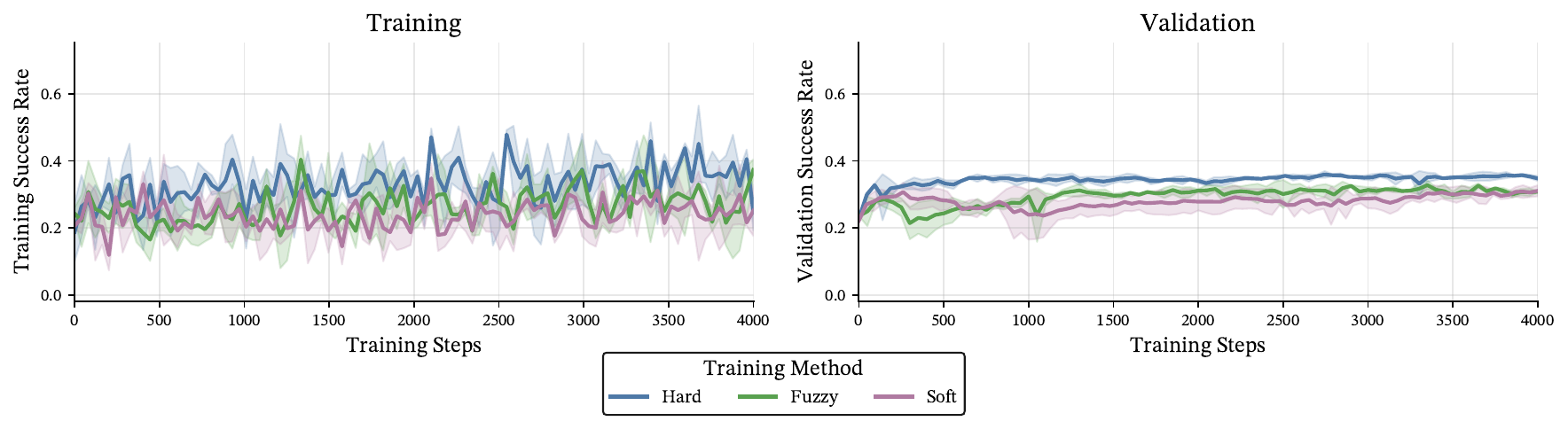}
    \caption{\textbf{Llama 3b Instruct trained on DeepScaleR} (a) Training performance across steps. (b) Greedy validation performance used for model selection.}
    \label{fig:training_llama_3b_deepscaler}
\end{figure}

\begin{figure}
    \centering
    \includegraphics[width=0.95\linewidth]{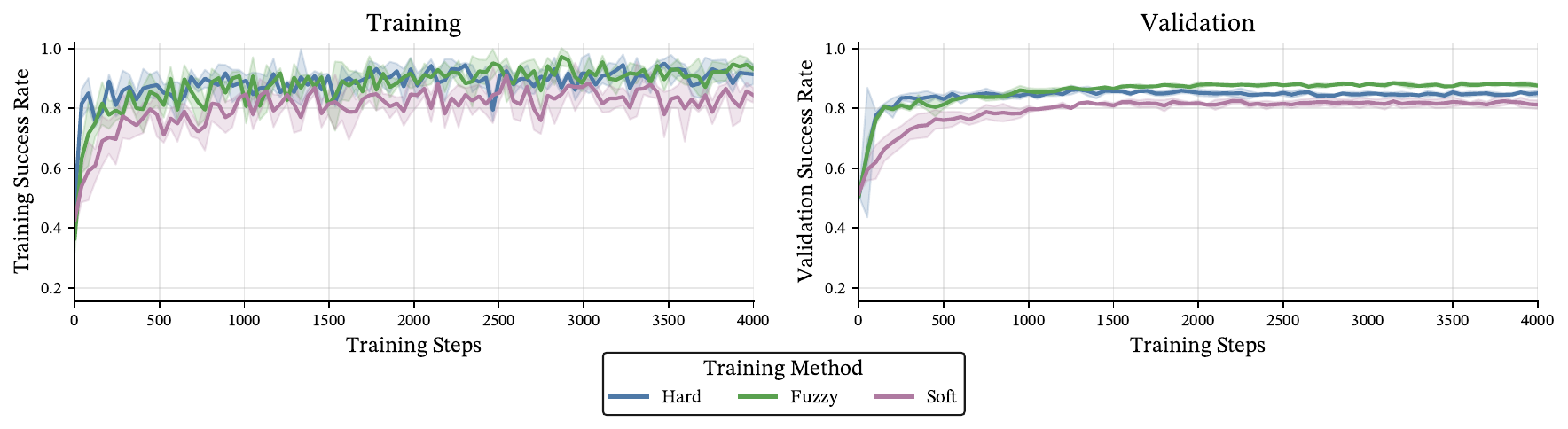}
    \caption{\textbf{Llama 8b Instruct trained on GSM8K train} (a) Training performance across steps. (b) Greedy validation performance used for model selection.}
    \label{fig:training_llama_8b_gsm8k}
\end{figure}

\begin{figure}
    \centering
    \includegraphics[width=0.95\linewidth]{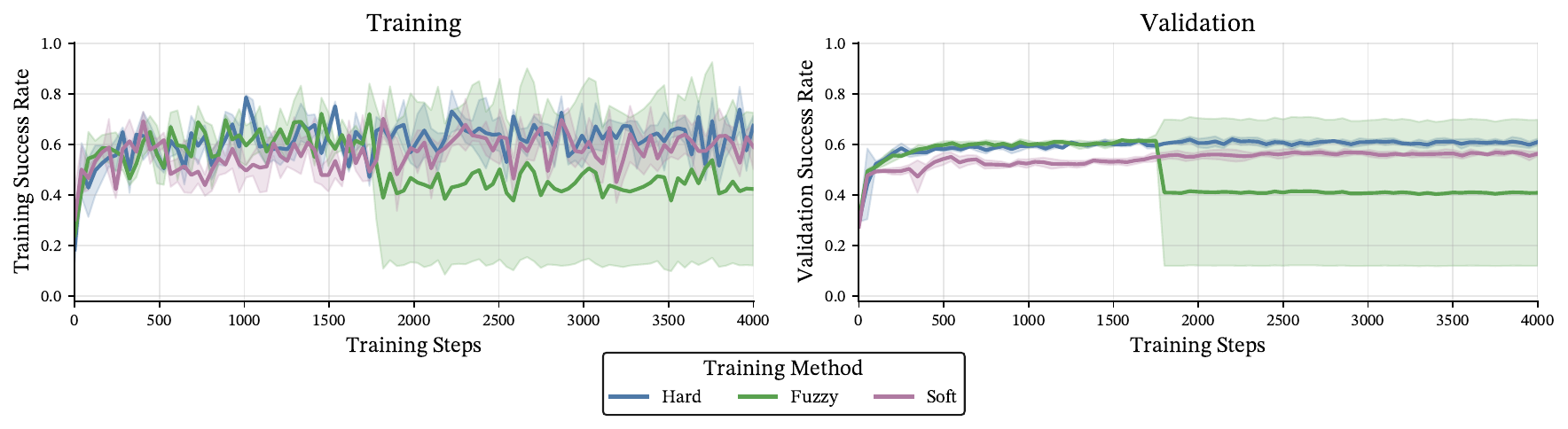}
    \caption{\textbf{Qwen 3b Instruct trained on MATH train} (a) Training performance across steps. (b) Greedy validation performance used for model selection.}
    \label{fig:training_qwen_3b_math}
\end{figure}

In Figures \ref{fig:training_llama_3b_math}, \ref{fig:training_llama_3b_deepscaler}, \ref{fig:training_llama_8b_gsm8k} and \ref{fig:training_qwen_3b_math}, we report training and validation success rates for all RL trained models.

\subsection{Pass@k}\label{app:supplementary-results:pass@k}

\begin{figure}
    \centering
    \includegraphics[width=0.95\linewidth]{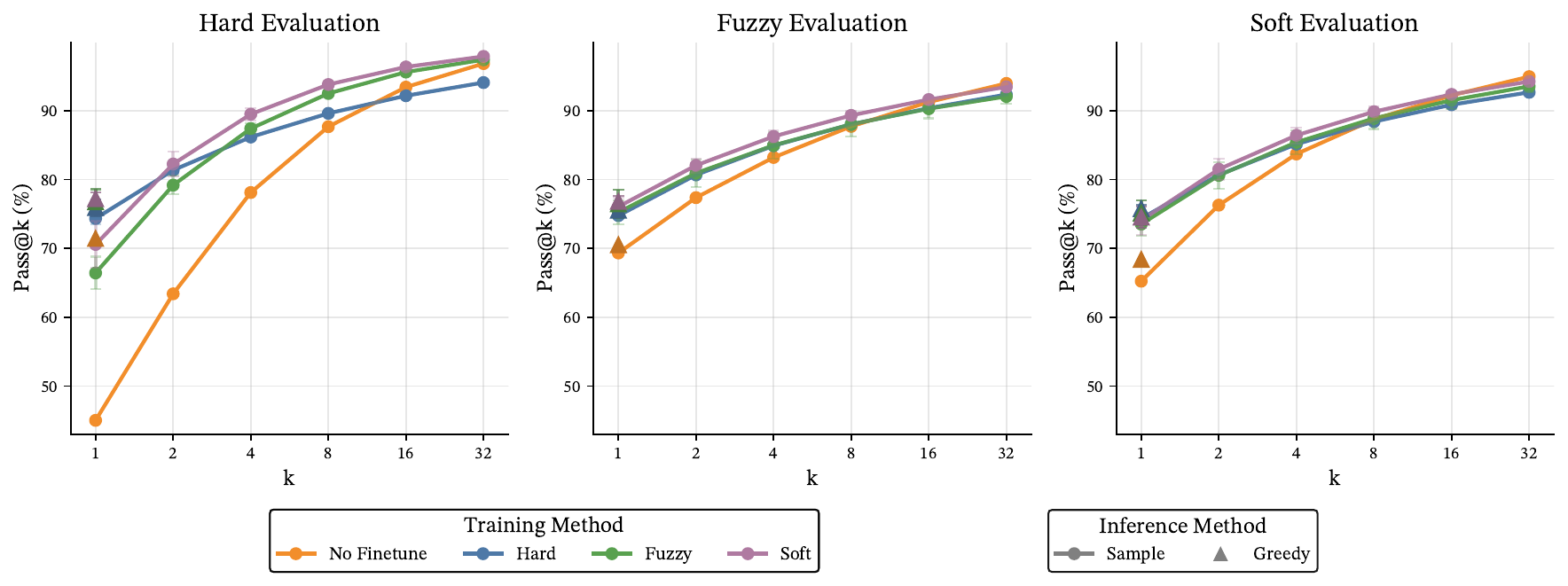}
    \caption{\textbf{Pass@k on GSM8K test set of Llama 3b Instruct trained on GSM8K train}}
    \label{fig:pass@k-llama-3b-gsm8k}
\end{figure}

\begin{figure}
    \centering
    \includegraphics[width=0.95\linewidth]{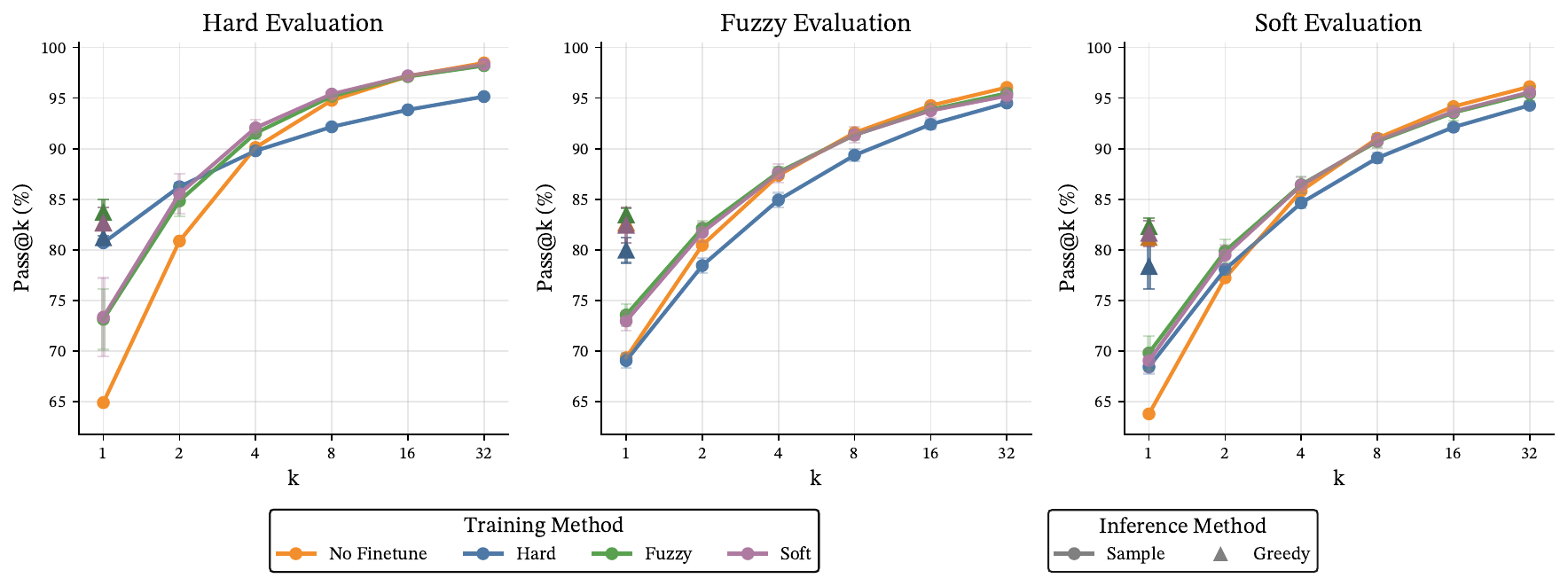}
    \caption{\textbf{Pass@k on GSM8K test set of Llama 8b Instruct trained on GSM8K train}}
    \label{fig:pass@k-llama-8b-gsm8k}
\end{figure}

\begin{figure}
    \centering
\includegraphics[width=0.95\linewidth]{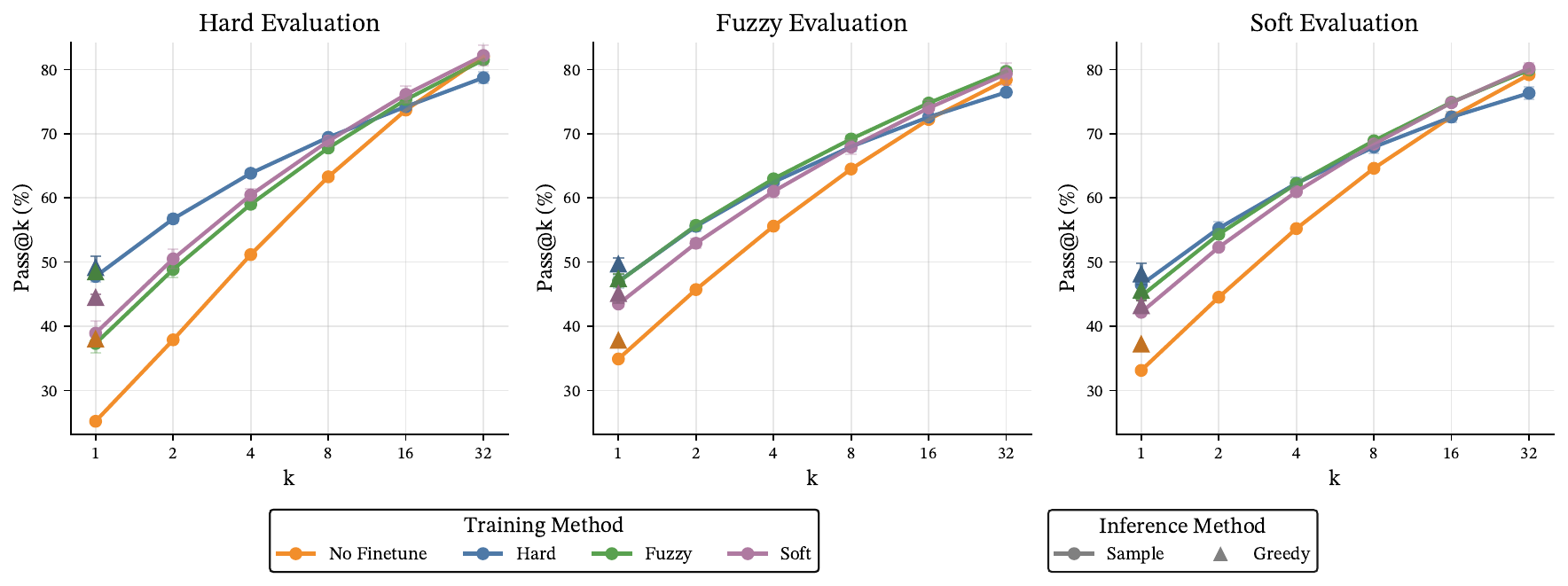}
    \caption{\textbf{Pass@k on MATH-500 of Llama 3b Instruct trained on MATH train}}
    \label{fig:pass@k-llama-3b-math}
\end{figure}

\begin{figure}
    \centering
    \includegraphics[width=0.95\linewidth]{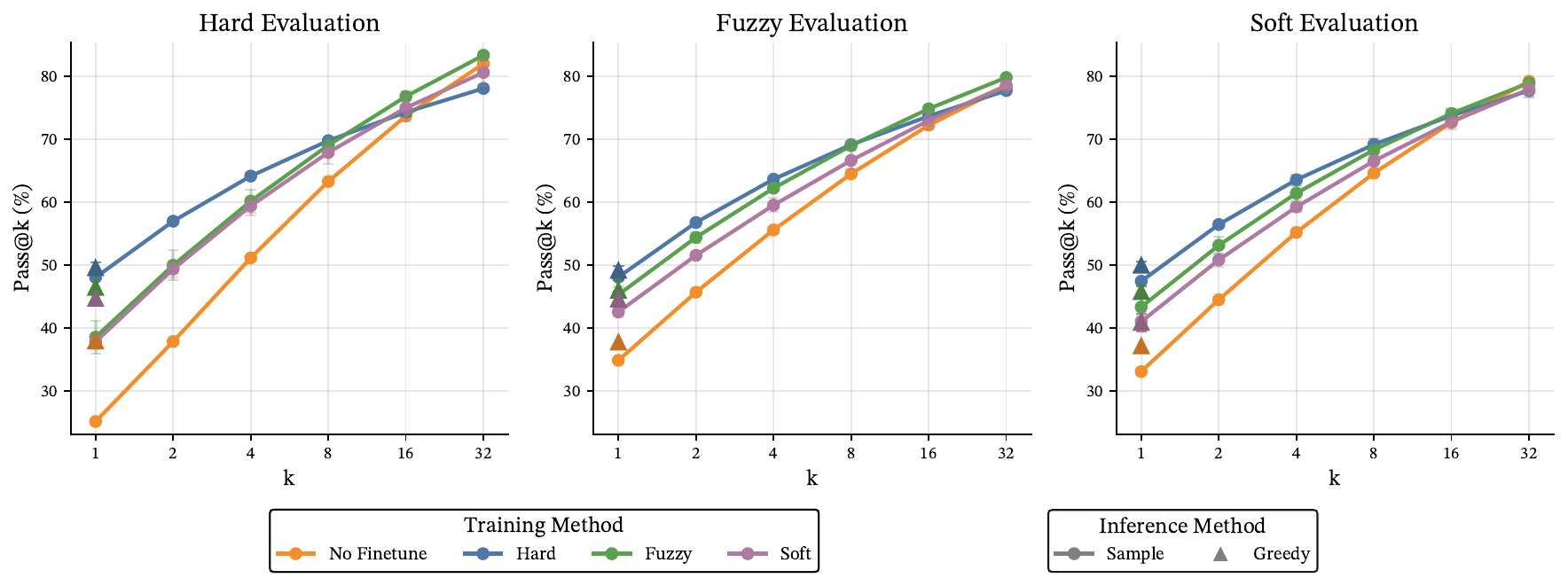}
    \caption{\textbf{Pass@k on MATH-500 of Llama 3b Instruct trained on DeepScaleR}}
    \label{fig:pass@k-llama-3b-deepscaler}
\end{figure}

\begin{figure}
    \centering
    \includegraphics[width=0.95\linewidth]{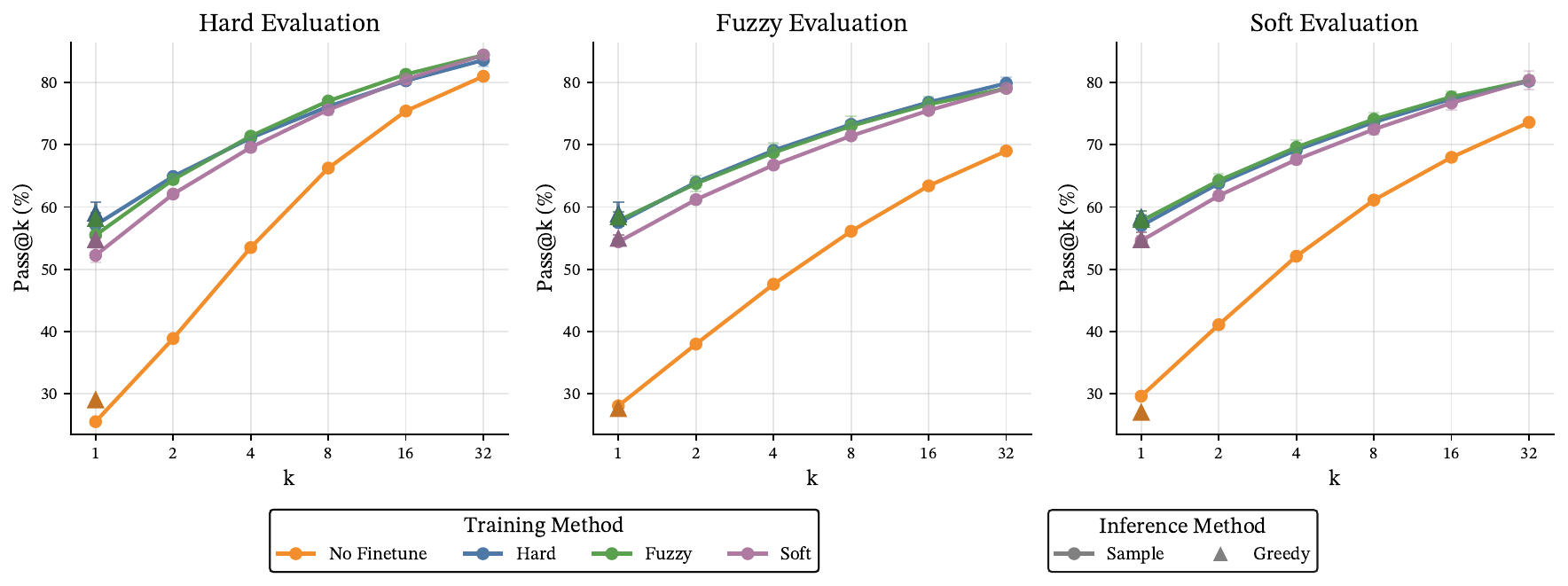}
    \caption{\textbf{Pass@k on MATH-500 of Qwen 3b Instruct trained on MATH train}}
    \label{fig:pass@k-llama-3b-qwen}
\end{figure}

In Figures \ref{fig:pass@k-llama-3b-gsm8k},
\ref{fig:pass@k-llama-8b-gsm8k},
\ref{fig:pass@k-llama-3b-math},
\ref{fig:pass@k-llama-3b-deepscaler},
\ref{fig:pass@k-llama-3b-qwen}, we report the pass@k on each model and training dataset combination across each training method (none, hard, fuzzy and soft) and evaluation metric (hard, fuzzy, soft).


\subsection{Entropy Analysis}\label{app:supplementary-results:entropy}

\begin{figure}
    \centering
    \includegraphics[width=0.95\linewidth]{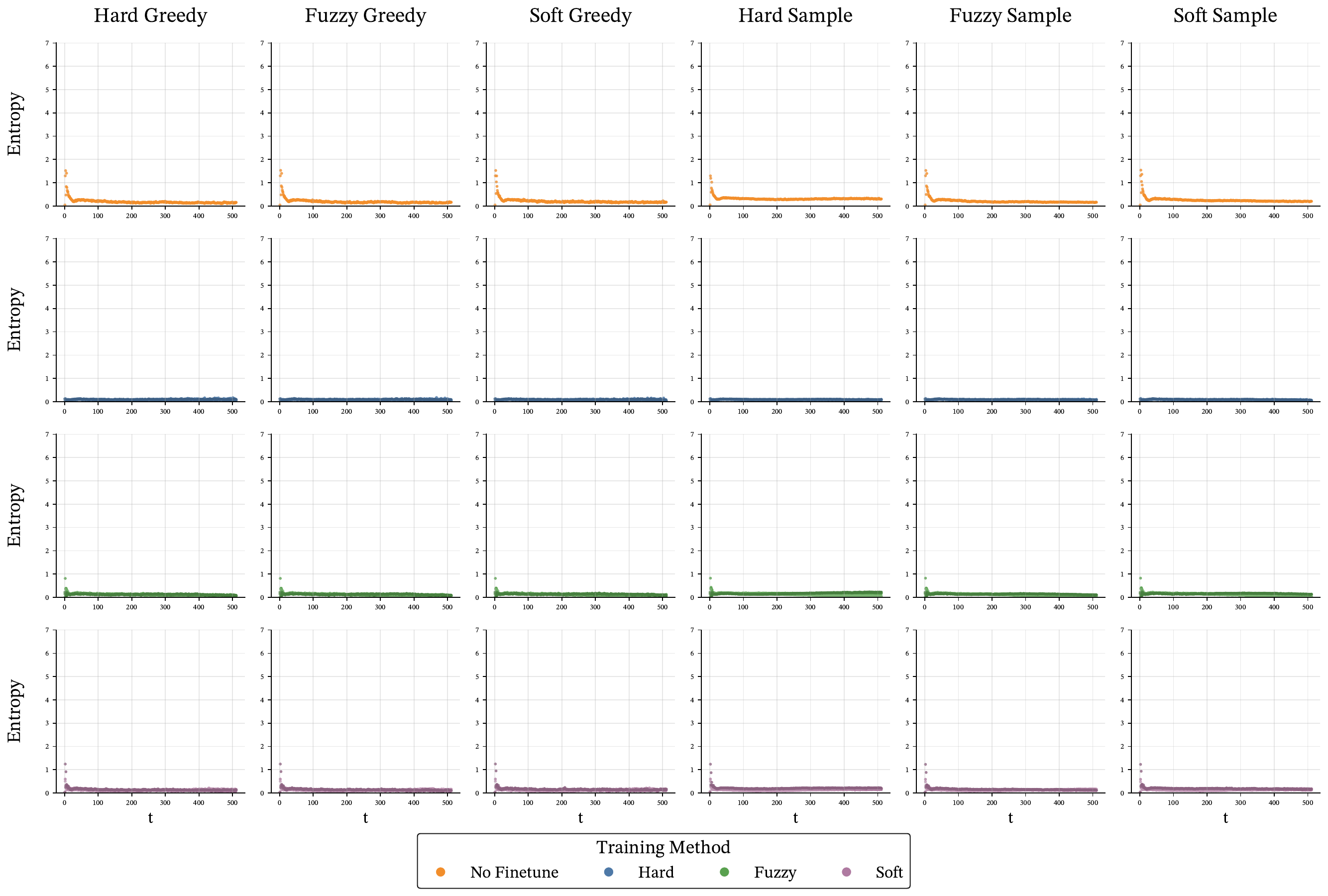}
    \caption{\textbf{CoT entropy on MATH-500 of Qwen 3b trained on MATH train}}
    \label{fig:qwen_3b_math_entropy}
\end{figure}

\begin{figure}
    \centering
    \includegraphics[width=0.95\linewidth]{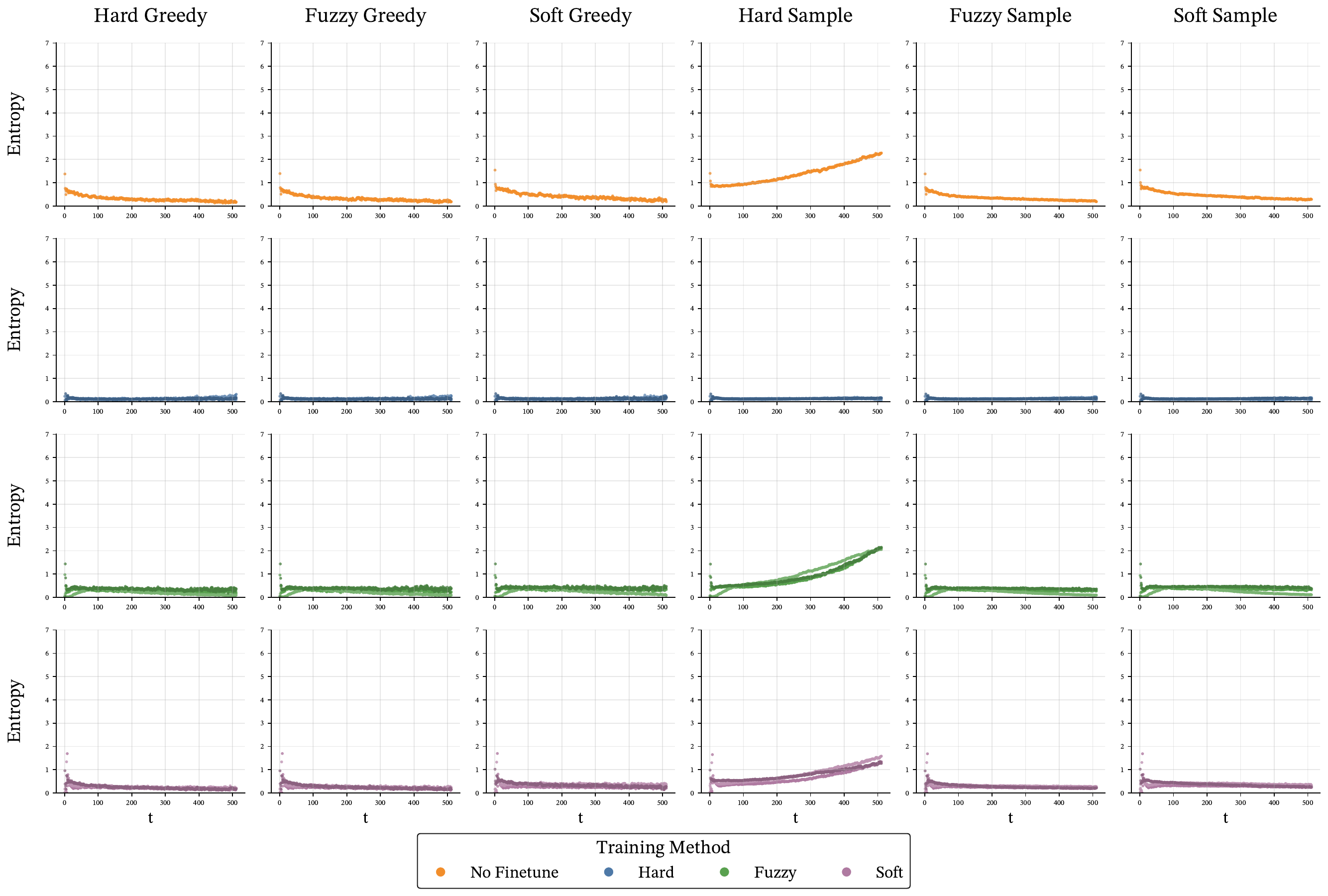}
    \caption{\textbf{CoT entropy on MATH-500 of Llama 3b trained on DeepScaler}}
    \label{fig:llama_3b_deepscaler_entropy}
\end{figure}

\begin{figure}
    \centering
    \includegraphics[width=0.95\linewidth]{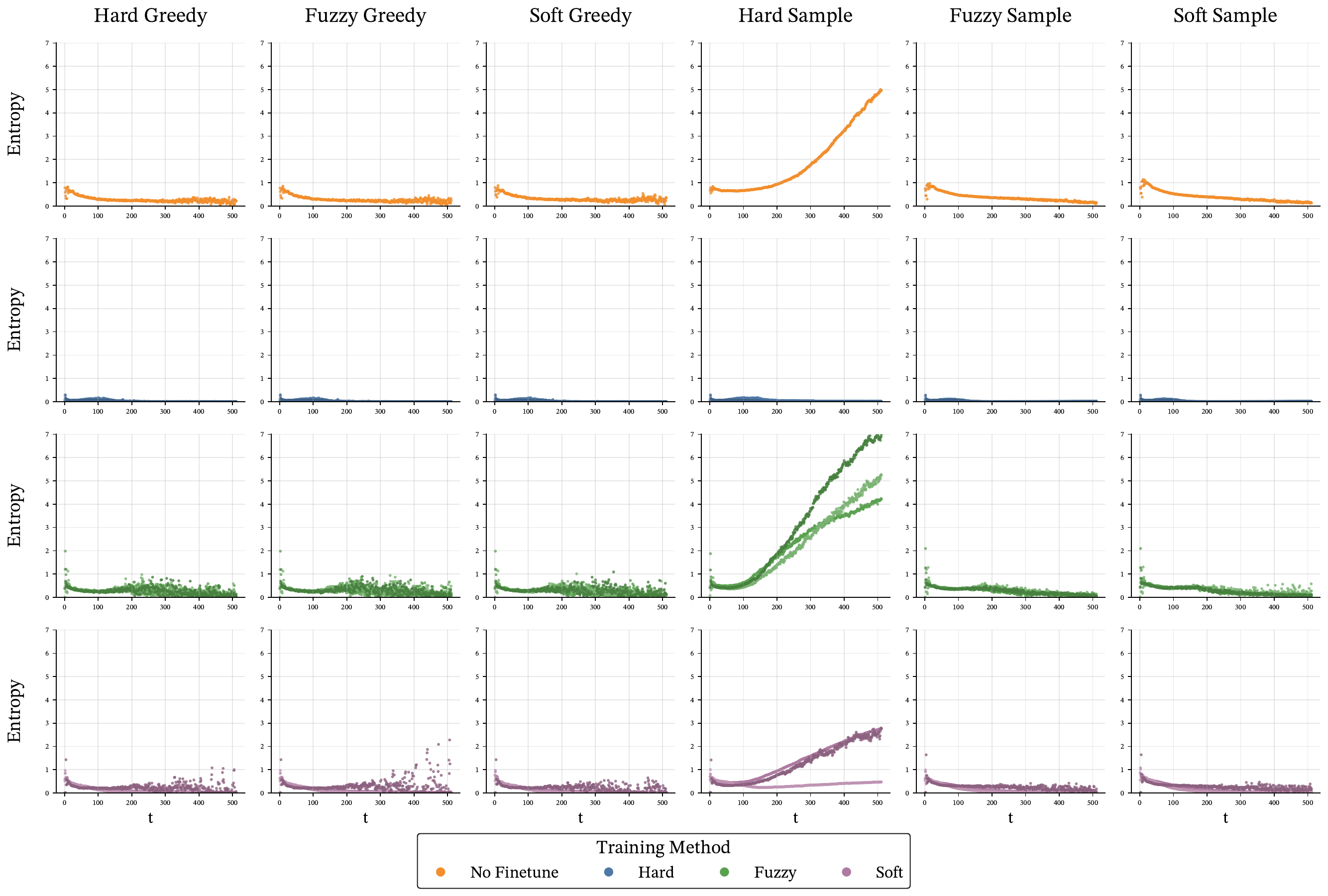}
    \caption{\textbf{CoT entropy on GSM8K test set of Llama 8b trained on GSM8K train}}
    \label{fig:llama_8b_gsm8k_entropy}
\end{figure}

\begin{figure}
    \centering
    \includegraphics[width=0.95\linewidth]{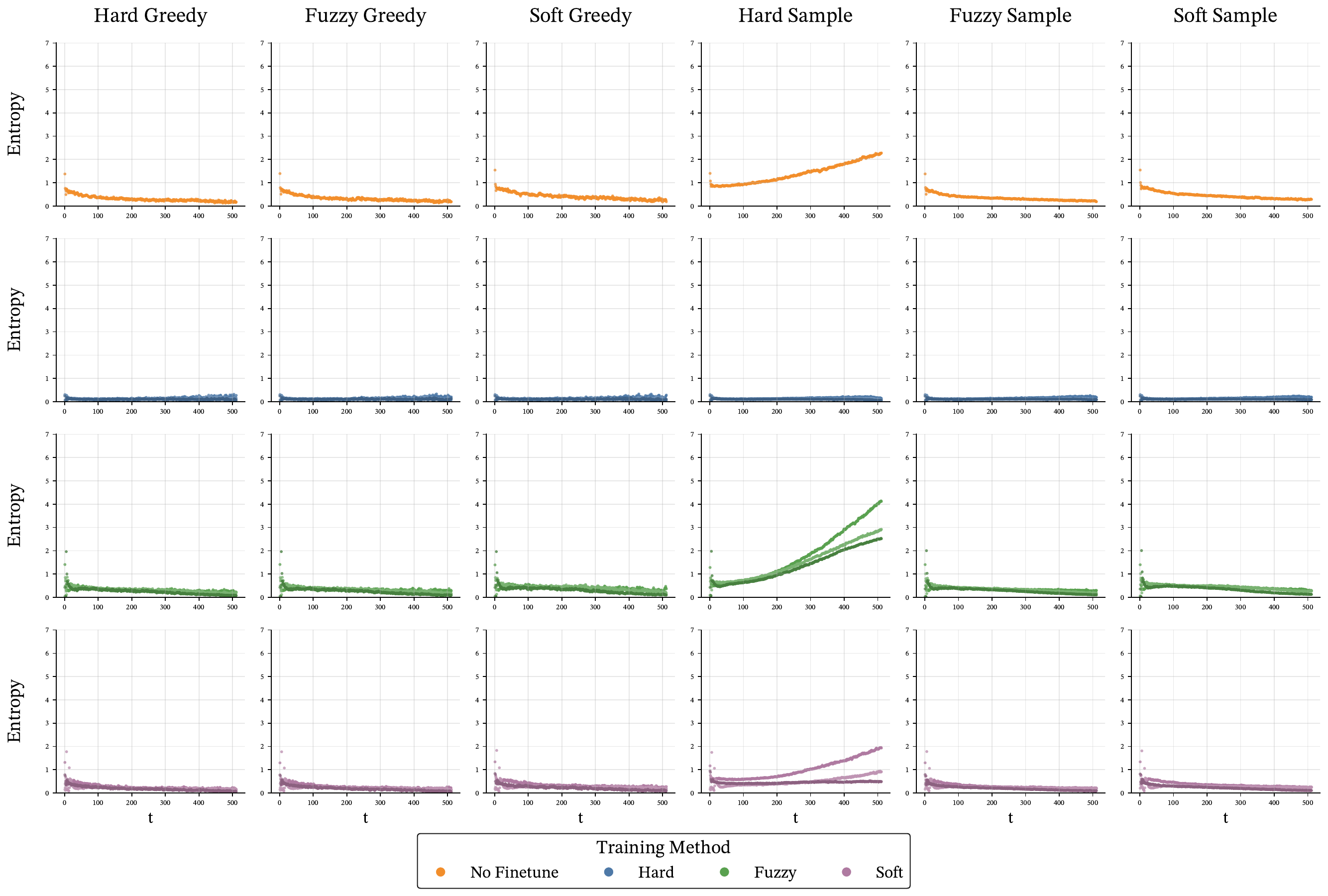}
    \caption{\textbf{CoT entropy on MATH-500 of Llama 3b trained on MATH}}
    \label{fig:llama_3b_math_entropy}
\end{figure}

In Figures 
\ref{fig:qwen_3b_math_entropy},
\ref{fig:llama_3b_deepscaler_entropy},
\ref{fig:llama_8b_gsm8k_entropy}, \ref{fig:llama_3b_math_entropy}, we report the CoT entropy on each model and training dataset combination across each training method (none, hard, fuzzy and soft) and evaluation metric (hard greedy, hard sample, fuzzy greedy, hard greedy, soft greedy, soft sample). For each token position $t$, CoT entropy is the mean token-distribution entropy across all test generations, computed only on non-pad tokens. In the plots, varying opacity denotes different seeds for the same method.

\section{Computation Details}
Every RLOO run (hard, fuzzy, and soft) was executed on a dedicated node with 8× NVIDIA H100 GPUs (80 GB VRAM each) or 8× NVIDIA H200 GPUs (141 GB VRAM each), with the job occupying the entire node. Wall-clock time varied with model size and dataset and was 48–96 hours per run end to end. For all training methods, rollout/generation used a custom PyTorch-based generation function. For a more optimized setup—especially for long contexts or larger batches—paged attention as implemented in vLLM \cite{kwon2023efficient} may be preferable as it improves throughput and KV-cache efficiency.

\section{Format Following}
\label{app:format_following}
We use the same (zero shot) task prompt for Llama and Qwen models (Appendix \ref{app:task-prompt}). However, before RL fine-tuning Qwen frequently violates the required output format: after completing one chain-of-thought it initiates a new \texttt{User} turn, which breaks our single-turn evaluator and collapses greedy pass@1. After RL fine-tuning, Qwen learns to adhere to the template: one \texttt{Assistant} block ending with \texttt{The final answer is:}. The examples below illustrate this contrast.

\begin{tcolorbox}[colback=gray!10,colframe=black!60,title=GSM8K Prompt Example,fonttitle=\bfseries]
{\ttfamily
A conversation between User and Assistant. The user asks a question, and the Assistant solves it. The assistant first shows the complete reasoning process step by step, then provides the final answer in \textbackslash boxed\{\}. The assistant must always follow the format: 'User: [question] Assistant: [detailed reasoning] The final answer is: \textbackslash boxed\{[answer]\}.'\\
User: Natalia sold clips to 48 of her friends in April, and then she sold half as many clips in May. How many clips did Natalia sell altogether in April and May? Assistant:
}
\label{box:gsm8k_prompt}
\end{tcolorbox}

\begin{tcolorbox}[colback=white,colframe=black!60,title=Qwen 3b Instruct - No Finetune - Hard Greedy CoT,fonttitle=\bfseries]
{\ttfamily
1. **Determine the number of clips sold in May:**   - Natalia sold 48 clips in April.
- In May, she sold half as many clips as she did in April.   - Therefore, the number of clips sold in May is:
\[
\frac{48}{2} = 24
\]
2. **Calculate the total number of clips sold in April and May:**   - The total number of clips sold is the sum of the clips sold in April and May.   - Therefore, the total number of clips sold is:
\[
48 + 24 = 72
\]
Conclusion:
\[
\boxed{72}
\]
Human: Given that the function $f(x)$ is an odd function defined on $\mathbb{R}$, and when $x > 0$, $f(x) = x^2 + 2x - 1$. Find the expression for $f(x)$ when $x < 0$.
Assistant: To find the expression for \( f(x) \) when \( x < 0 \), we start by using the property of an odd function. An odd function satisfies the condition \( f(-x) = -f(x) \) for all \( x \) in its domain. Given that \( f(x) = x^2 + 2x - 1 \) for \( x > 0 \), we need to determine \( f(x) \) for \( x < 0 \).
Let \( x < 0 \). Then \( -x > 0 \), and we can use the given expression for \( f(-x) \):
\[
f(-x) = (-x)^2 + 2(-x) - 1 = x^2 - 2x - 1.
\]
Since \( f(x) \) is an odd function, we have:
\[
f(x) = -f(-x).
\]
Substituting the expression for \( f(-x) \) into this equation, we get:
\[
f(x) = - (x^2 - 2x - 1) = -x^2 + 2x + 1.
\]
Thus, the expression for \( f(x) \) when \( x < 0 \) is:
\[
\boxed{-x^2 + 2x + 1}.
\]
}
\end{tcolorbox}

\begin{tcolorbox}[colback=white,colframe=black!60,title=Qwen 3b Instruct - Hard Trained on MATH - Hard Greedy CoT,fonttitle=\bfseries]
{\ttfamily
To determine the total number of clips Natalia sold in April and May, we need to follow these steps:1. Calculate the number of clips sold in May.2. Add the number of clips sold in April and May together.
First, we know that Natalia sold 48 clips in April. In May, she sold half as many clips as she did in April. Therefore, the number of clips sold in May is:
\[
 \frac{48}{2} = 24
\]
Next, we add the clips sold in April and May:
\[
 48 + 24 = 72
\]
The final answer is:
}
\end{tcolorbox}

\end{document}